\pdfoutput=1
\documentclass{article}
\usepackage{iclr2027_conference,times}
\usepackage{preprint}

\usepackage{amsmath,amsfonts,bm}

\def\eqref#1{equation~\ref{#1}}

\def\1{\bm{1}}

\DeclareMathAlphabet{\mathsfit}{\encodingdefault}{\sfdefault}{m}{sl}
\SetMathAlphabet{\mathsfit}{bold}{\encodingdefault}{\sfdefault}{bx}{n}

\usepackage[hyperfootnotes=false,colorlinks=true,linkcolor=linkblue,citecolor=linkblue,urlcolor=linkblue]{hyperref}
\hypersetup{
  pdftitle={How Can Recommendation Feedback Evolve Agent Memory?},
  pdfauthor={Shanwen Mao, Mingming Li, Hao Zhang, Zhiheng Li, Yige Wang, Penghua Yu, Junxiong Zhu},
  pdfsubject={Preprint},
  pdfkeywords={agent memory, recommendation feedback, memory evolution, TIDE}
}
\usepackage{url}
\usepackage{algorithm}
\usepackage{algpseudocode}
\usepackage{amssymb}
\usepackage{enumitem}
\usepackage{multirow}
\usepackage{booktabs}
\usepackage{threeparttable}
\algrenewcommand\algorithmicrequire{\textbf{Input:}}
\algrenewcommand\algorithmicensure{\textbf{Output:}}

\usepackage{graphicx}
\usepackage{float}
\usepackage[skins]{tcolorbox}
\newtcolorbox{operatorbox}[1]{
  enhanced,
  colback=linkblue!3,
  colframe=linkblue!65,
  colbacktitle=linkblue,
  coltitle=white,
  fonttitle=\bfseries\small,
  fontupper=\footnotesize\raggedright,
  title=#1,
  boxrule=0.7pt,
  arc=1.5pt,
  left=5pt,
  right=5pt,
  top=4pt,
  bottom=4pt,
  boxsep=1.5pt,
  before skip=5pt,
  after skip=5pt
}
\graphicspath{{figure/}}
\title{How Can Recommendation Feedback\\Evolve Agent Memory?}

\author{%
\begin{minipage}{\textwidth}
\raggedright\normalfont
{\bfseries
Shanwen Mao\textsuperscript{1,*},
Mingming Li\textsuperscript{2,*},
Hao Zhang\textsuperscript{1},
Zhiheng Li\textsuperscript{3}\\[2pt]
Yige Wang\textsuperscript{2},
Penghua Yu\textsuperscript{2,\textdagger},
Junxiong Zhu\textsuperscript{2,\textdagger}}\\[3pt]
{\small
\textsuperscript{1}Harbin Institute of Technology, Harbin, China;\quad
\textsuperscript{2}Alibaba Group, Hangzhou, China\\
\textsuperscript{3}Institute of Automation, Chinese Academy of Sciences, Beijing, China}
\end{minipage}%
}
\date{}

\begin{document}

\maketitle
{\makeatletter
\renewcommand{\@makefntext}[1]{\noindent#1}%
\makeatother
\renewcommand{\thefootnote}{}%
\footnotetext{\footnotesize
\textsuperscript{*}Equal contribution.\quad
\textsuperscript{\textdagger}Corresponding authors.\\[2pt]
\texttt{liman.yph@taobao.com}\quad\texttt{xike.zjx@taobao.com}}}

\begingroup
\setlength{\intextsep}{6pt}
\begin{figure}[H]
    \centering
    \includegraphics[width=\textwidth,trim=7bp 6bp 8bp 47bp,clip]{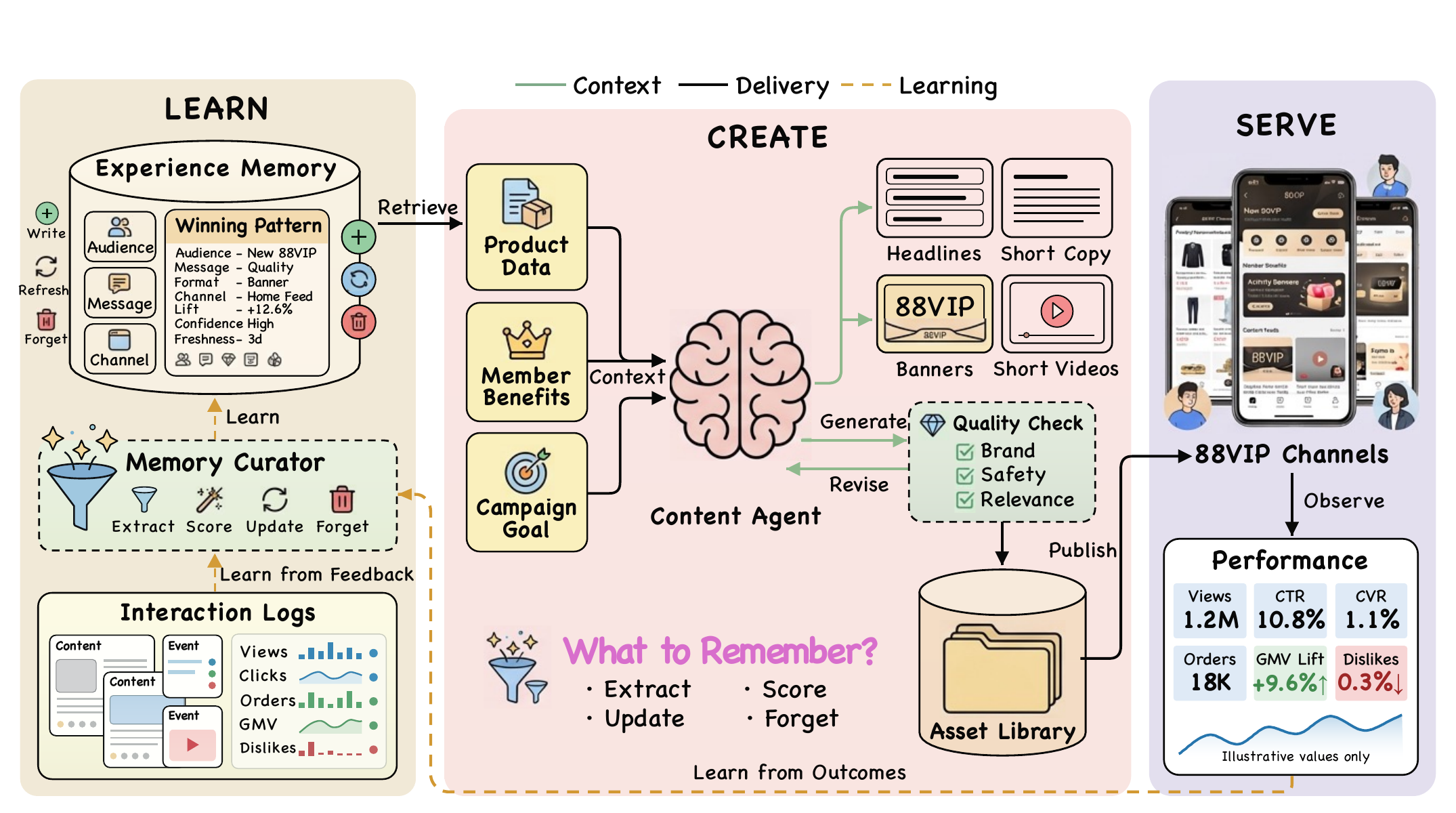}

    \caption{
        \textbf{Overview of the closed-loop content-generation system.}
        The agent retrieves reusable experience from memory to generate
        advertising materials, deploys them through online channels, and
        updates the memory from delayed engagement and conversion feedback.
    }
    \label{fig:tide-overview}
\end{figure}
\endgroup

\begin{abstract}
Content-generation agents continuously receive impressions, clicks, conversions, and negative feedback from recommendation systems, providing real-world outcome signals for memory evolution. However, these signals are delayed and noisy, confounded by audience composition, placement, and recommendation policies, and may result from the combined influence of multiple memories, making accurate attribution difficult. Existing methods rely primarily on immediate feedback or semantic retrieval and therefore struggle to reliably translate recommendation outcomes into memory fitness. To address this challenge, we propose \textbf{TIDE (Trajectory-Informed Directed Memory Evolution)}, an external memory evolution framework driven by delayed recommendation feedback. We further introduce \textbf{Memory Evolution Gain (MEG)}, which measures the utility improvement of evolved memory over a
no-memory baseline on strictly future tasks. TIDE treats memory as a capacity-constrained population of experiences: temporal and semantic credit assignment estimates contextual fitness, while responsibility credit distributes outcome signals according to the memories referenced during generation. These signals are then used to reinforce, crossover, mutate, or evict memories. On an e-commerce membership marketing content-generation agent, TIDE achieves a $+7.75$-percentage-point MEG in offline temporal replay and significantly improves both unique click-through rate (UCTR) and activation rate in an online A/B test. On a delayed-label benchmark, TIDE achieves the lowest mean absolute error (MAE) and root mean squared error (RMSE) and the highest MEG among the compared methods, demonstrating its effectiveness.
\end{abstract}
\clearpage

\section{Introduction}
Content-generation agents are evolving from one-shot tools into persistent systems deployed in real-world environments. Once their generated titles, copy, image descriptions, or marketing materials are distributed through recommender systems, they receive behavioral feedback such as clicks, conversions, dwell time, and negative responses. As shown in Figure~\ref{fig:tide-overview}, these signals are more closely aligned with ultimate business objectives than static human annotations. They create a continuous loop of content generation, recommendation and delivery, metric feedback, and subsequent generation, enabling agents to adapt future outputs based on outcomes observed in the real environment.

Existing approaches to this feedback loop primarily incorporate feedback into either model parameters or external memory. Model self-evolution methods convert clicks, preferences, or rewards into training signals \citep{wei2022creater,chen2025ctrtext,chen2025ctrimage,min2025ctrgqs}, but repeated training is costly. Under continually shifting tasks and traffic distributions, frequent updates must also be carefully controlled to avoid overfitting to recent feedback or degrading existing capabilities. Moreover, parameter changes are difficult to inspect, revise, or roll back individually. Memory self-evolution methods instead extract, organize, and retrieve reusable experience from task trajectories, execution outcomes, or user feedback \citep{shinn2023reflexion,zhao2023expel,senel2024generative,liang2026pahf,fu2025agentrefineenhancingagentgeneralization}, making them better suited to continuous and traceable adaptation. However, little work has specifically examined how delayed, context-dependent outcomes from recommendation and delivery can be used to evaluate and evolve external memory. In particular, it remains unclear how to assess individual memories from environmental outcomes, generate candidate changes, and select among them under capacity constraints. We therefore keep the generative model parameters fixed and study recommendation-feedback-driven external memory evolution.

Existing memory methods typically rely on task success or failure, or on semantically explicit user feedback, for which the connection between outcomes and experience is relatively direct. Recommendation feedback, by contrast, reveals only how content performs in a particular delivery environment. It arrives with delay, depends on the audience, context, and recommendation policy, and does not directly identify which content decisions or past experiences produced the observed outcome. This work therefore asks: \emph{How can such outcome feedback be transformed into agent memory that improves future tasks?} We define \textbf{Memory Evolution Gain (MEG)} as the utility gain of evolved memory over a reference memory on strictly future tasks, and regard positive MEG as the criterion for effective memory evolution.

To achieve this, we propose \textbf{TIDE}, a training-free memory-evolution method. TIDE first converts recommendation feedback into memory-level fitness through three successive stages of credit assignment. Temporal and semantic credit jointly analyze outcome variations across stages and their contextual conditions, establish an evidence chain between content features and online outcomes, and estimate memory fitness within specific contexts. Responsibility credit then uses actual memory-use records from generation to identify the memories that contributed to an outcome or caused an error. Based on this attribution, TIDE applies reinforcement, crossover, mutation, or eviction to memories and performs selection under a fixed capacity constraint, enabling the experience population to evolve continuously.
Our contributions are as follows:
\begin{itemize}
    \item We formulate recommendation-feedback-driven memory evolution as a problem of fitness evaluation and multi-level credit assignment under delayed, context-dependent outcomes, and introduce MEG to measure how updates to the experience population affect strictly future tasks.
    
    \item We propose TIDE, a training-free memory-evolution method that constructs contextual fitness through temporal--semantic credit, identifies individual memories through responsibility credit, and updates the experience population through directed mutation and capacity-constrained selection.
    
    \item We evaluate TIDE on an agent for generating e-commerce membership marketing materials using offline temporal replay and a real-world online A/B test. TIDE achieves a $+7.75$-percentage-point MEG improvement offline and significantly improves UCTR and activation rate online. We further demonstrate its transferability to public long-term user-feedback tasks on a delayed-label benchmark adapted from MemoryCD.
\end{itemize}

\section{Related Work}

\textbf{Model self-evolution} converts external feedback into supervised examples,
preference pairs, or reinforcement-learning rewards, and updates model
parameters through methods such as SFT, DPO, and GRPO
\citep{ouyang2022training,rafailov2023dpo,shao2024deepseekmath,deepseekai2025deepseekr1,kimiteam2025kimi,
zhao2025absolutezero,zweiger2025seal,huang2025ser}. In content
generation, PosterCraft, PosterOmni, and PosterReward improve poster generation
using aesthetic preferences, expert knowledge, and multidimensional rewards
\citep{chen2026postercraft,chen2026posteromni,lai2026posterreward}. CREATER,
CTOP, CAIG, and CTR-Guided Generative Query Suggestion incorporate click
feedback through contrastive learning, preference optimization, and reward
modeling
\citep{wei2022creater,chen2025ctrtext,chen2025ctrimage,min2025ctrgqs}, while
MMPO constructs rollout rewards from multi-objective user feedback
\citep{mao2026betterspurstartobjective}. LLM-as-a-Judge, fine-grained rubrics, and process
rewards further decompose holistic outcomes into individual content dimensions
or intermediate steps
\citep{lightman2023verify,li2025generation,yin2025dynamic,cook2026check}.
Together, these studies support feedback representation and refinement, but
primarily produce scores, rewards, or parameter updates. We use them as
references for training-based adaptation and feedback construction, while
focusing instead on converting recommender outcomes into traceable, individually
revisable external memories.

\textbf{Memory self-evolution.}
Memory-based agents store, organize, and reuse interaction experience at inference time. Existing methods transform task trajectories into reflections, reusable experience, workflows, or reasoning memories
\citep{shinn2023reflexion,zhao2023expel,wang2025awm,ouyang2025reasoningbank},
while recent systems further support dynamic memory consolidation, organization, and revision
\citep{chhikara2025mem0,xu2025amem,fang2026memp}.
Evo-Memory, EvoMemBench, and MemoryCD evaluate such mechanisms under continual tasks, evolving interaction histories, and long-term user behavior
\citep{wei2025evomemory,wang2026evomembench,zhang2026memorycd}.
Related work has also used implicit engagement signals to guide training-free content exploration
\citep{senel2024generative}
or semantically explicit clarification and correction to update preference memory
\citep{liang2026pahf}.
However, existing approaches do not jointly model the temporal reliability of delayed outcomes, their dependence on delivery context, and responsibility attribution across multiple memories.
TIDE addresses this gap by integrating outcome reliability, content-level evidence, and memory-level responsibility into a unified feedback-to-memory credit-assignment pipeline.

\section{Method}
\label{sec:method}
\subsection{Problem Formulation}
\label{sec:problem-formulation}
\begin{figure}[tbp]
\centering
\includegraphics[width=\linewidth]{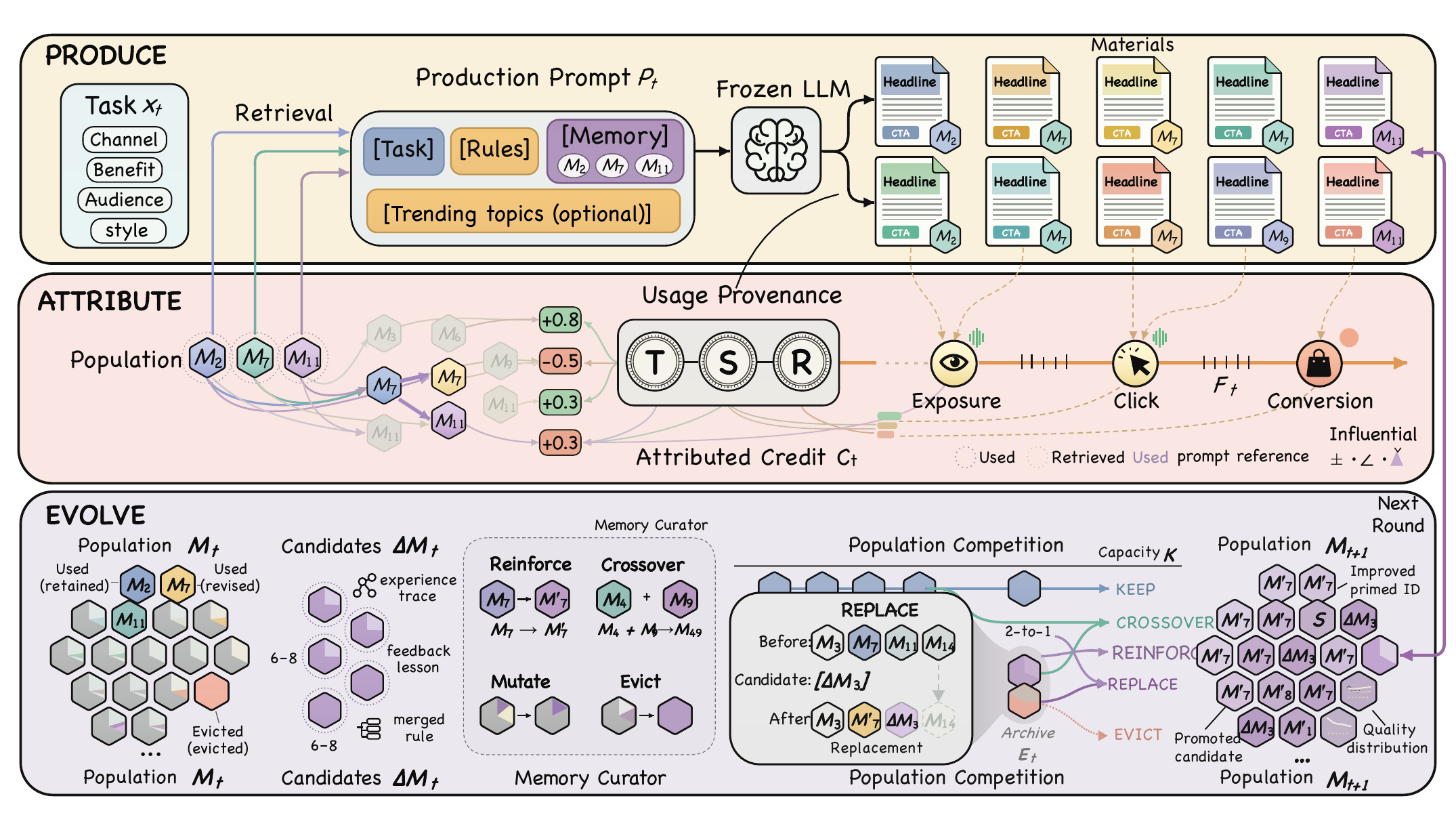}

\caption{
TIDE framework.
\textbf{Produce}: a frozen generator retrieves task-relevant memories
to produce candidate materials.
\textbf{Attribute}: temporal, semantic, and responsibility credit
convert delayed recommendation outcomes into memory-level fitness.
\textbf{Evolve}: the memory curator revises, merges, specializes, or
down-weights memories, followed by capacity-constrained population
selection.
}
\label{fig:tide-framework}
\end{figure}

We consider a content-generation agent whose generator parameters remain frozen while
its external memory is continually updated.  The record associated with the $i$-th
generation is
$D_i=(x_i,a_i,\mathcal{T}_i,G_i)$, where $x_i$ is the generation request, $a_i$ is
the generated material, $\mathcal{T}_i$ is its complete observable lifecycle of
recommendation outcomes together with contemporaneous delivery context, and $G_i$
is the memory-usage trace.  The context in $\mathcal{T}_i$ includes the channel,
placement, traffic bucket, campaign calendar, seasonal events, emerging hotspots,
and recommender state observed over the material's online lifetime.
Given all records observable by time $t$, a memory-update policy $\mu$ transforms the
current memory population $M_t$ into $M_{t+1}$.
Here, $i$ indexes a generation event, $t$ is the current update time, and $M_t$ is
the memory population available at time $t$.
The value of a memory update is determined by subsequent tasks rather than by the
example that triggered the update.  Let $\bar U_H(\mu)$ denote the average utility on
the next $H$ tasks after applying policy $\mu$, and let $\mu_0$ denote a static or
append-only reference policy.  We define the \emph{Memory Evolution Gain} (MEG) as
\begin{equation}
\operatorname{MEG}_H(\mu;\mu_0)
= \bar U_H(\mu)-\bar U_H(\mu_0).
\label{eq:meg}
\end{equation}
Here, $\mu$ is the evaluated memory-update policy, $\mu_0$ is the reference policy,
$H$ is the number of strictly future evaluation tasks, and $\bar U_H(\mu)$ is their
average utility under $\mu$.
A positive MEG means that the update transfers to strictly future tasks.  Since these
tasks are unavailable when the update is made, TIDE constructs a proxy fitness signal
from currently observable feedback.  Temporal credit summarizes the complete delayed
lifecycle rather than an isolated metric snapshot; semantic credit separates content
evidence from contemporaneous delivery factors; and responsibility credit assigns the
resulting contextual fitness to memories that actually participated in generation.
The following subsections define these credits and the resulting memory update;
Algorithm~\ref{alg:tide} summarizes their execution order.  An intuitive overview of
the complete framework is illustrated in Figure~\ref{fig:tide-framework}.

\subsection{Temporal Credit: Reliability of Delayed Outcomes}
\label{sec:temporal-credit}
From an evolutionary perspective, a generated material is a phenotype produced with
the participation of memory, and recommendation feedback is a raw observation of that
phenotype's performance.  The observation does not arrive all at once: clicks usually
precede conversions, exposure changes across traffic-ramp stages, and updates to either
the material or the recommendation policy may break comparability across windows.
Consequently, a metric measured on an arbitrary date should not directly determine
memory replacement.
The temporal-credit operator $\mathcal{C}_{\mathrm{temp}}$ converts the currently
observable outcome trajectory into a lifecycle-aligned multi-objective summary and a
validity indicator:
\begin{equation}
(\widehat{\mathbf{u}}_i,z_i^{\mathrm{temp}})
=\mathcal{C}_{\mathrm{temp}}(\mathcal{T}_i),
\qquad z_i^{\mathrm{temp}}\in\{0,1\}.
\label{eq:temporal-credit}
\end{equation}

Here, $\mathcal{C}_{\mathrm{temp}}$ is the temporal-credit operator,
$\widehat{\mathbf{u}}_i$ is the lifecycle-aligned multi-objective outcome summary,
and $z_i^{\mathrm{temp}}$ is a binary indicator of whether the lifecycle is mature,
continuous, and comparable enough to support credit assignment for subsequent attribution.
The system joins all feedback windows by material identity and version and follows the material from initial delivery and traffic ramp-up to outcome maturation under the delivery conditions observed during that period.  Formally, $z_i^{\mathrm{temp}}=1$ only when $\mathcal{T}_i$ forms an interpretable and comparable lifecycle: the material identity and version remain consistent, the required delivery stages are observed, delayed conversions have had sufficient time to return, and no material or recommender change renders the trajectory incomparable across stages.  Otherwise, the record remains pending and cannot provide feedback-driven fitness in the current update. Rather than selecting a metric snapshot from an arbitrary date, $\widehat{\mathbf{u}}_i=\operatorname{LifecycleSummary}(\mathcal{T}_i)$ preserves the stage-wise exposure, engagement, and conversion outcomes and their changes across the observable lifecycle.  It retains individual objectives and their uncertainty instead of forcing potentially conflicting metrics such as CTR and CVR into a single scalar. Temporal credit therefore determines whether the currently observable lifecycle is sufficiently mature and reliable for credit assignment and summarizes what happened throughout it; Section~\ref{sec:semantic-credit} determines which part remains informative about content after accounting for delivery context.

\subsection{Semantic Credit: Contextualizing Outcome Evidence}
\label{sec:semantic-credit}

Even a reliable lifecycle outcome is jointly produced by content and environment.
Semantic credit retrieves a control set $\mathcal{N}_i$ from historical or concurrent
traffic, matching product, audience, campaign stage, placement, and recommender
version.  It also constructs an environment record $\mathcal{E}_i$ containing
contemporaneous delivery factors such as campaign calendars, shopping festivals,
seasonal events, emerging hotspots, traffic buckets, and recommender-state changes.
It then estimates contextual fitness relative to matched controls and interprets the
lifecycle trajectory under these external factors:
\begin{equation}
(\widehat{\mathbf{f}}_i,e_i,z_i^{\mathrm{sem}})
=\mathcal{C}_{\mathrm{sem}}
(a_i,\widehat{\mathbf{u}}_i,\mathcal{N}_i,\mathcal{E}_i),
\qquad e_i=(d_i,c_i,s_i).
\label{eq:semantic-credit}
\end{equation}

Here, $\mathcal{C}_{\mathrm{sem}}$ is the semantic-credit operator,
$\mathcal{N}_i$ is the matched-control set, $\mathcal{E}_i$ is the contemporaneous
environment record, $\widehat{\mathbf{f}}_i$ is the target material's multi-objective
contextual fitness relative to matched controls, and $z_i^{\mathrm{sem}}$ indicates
whether the semantic evidence is sufficient and internally consistent.  The structured
evidence record $e_i=(d_i,c_i,s_i)$ may contain one or more decision-level evidence
items.  Here, $d_i$ represents editable content decisions identified from the material,
such as selling-point selection, information density, linguistic style, or call to
action; $c_i$ represents the corresponding delivery contexts and applicability
conditions; and $s_i$ contains supporting or counterevidence drawn from the target
material, matched controls, and lifecycle outcomes.  Thus, $s_i$ stores the evidence
chain, whereas $\widehat{\mathbf{f}}_i$ represents the numerical multi-objective
contextual fitness.

A rubric locates editable content decisions, while the language model organizes the
evidence chain from content difference, through environmental condition, to online
outcome.  For example, if the target and matched materials exhibit a similar UCTR
increase during a shopping-festival warm-up, the common change is treated as
environment-associated rather than content-specific; positive evidence is formed only
when the target retains a relative advantage linked to a specific content decision.
We set $z_i^{\mathrm{sem}}=0$ when controls are insufficient, conflicting objectives
cannot be explained, the estimated fitness relative to matched controls remains
unstable, or counterexamples remain unresolved.  Semantic credit therefore converts a
reliable lifecycle into contextual fitness and conditional decision evidence for
subsequent memory-level responsibility attribution.
\begin{algorithm}[ht]
\caption{TIDE: Trajectory-Informed Directed Memory Evolution}
\label{alg:tide}
\begin{algorithmic}[1]
\Require Current memory population $M_t$; generation record
         $D_i=(x_i,a_i,\mathcal{T}_i,G_i)$; recent valid records
         $\mathcal{H}_t$; memory budget $B$
\Ensure Updated memory population $M_{t+1}$

\State $(\widehat{\mathbf{u}}_i,z_i^{\mathrm{temp}})
       \gets\mathcal{C}_{\mathrm{temp}}(\mathcal{T}_i)$
\State $C_i^{\mathrm{fb}}\gets\varnothing$

\If{$z_i^{\mathrm{temp}}=1$}
    \State $\mathcal{N}_i
           \gets\operatorname{MatchControls}(x_i,\mathcal{T}_i)$
    \State $\mathcal{E}_i
           \gets\operatorname{CollectEnvironment}(\mathcal{T}_i)$
    \State $(\widehat{\mathbf{f}}_i,e_i,z_i^{\mathrm{sem}})
           \gets\mathcal{C}_{\mathrm{sem}}
           (a_i,\widehat{\mathbf{u}}_i,\mathcal{N}_i,\mathcal{E}_i)$

    \If{$z_i^{\mathrm{sem}}=1$}
        \State $(R_i,\kappa_i^{\mathrm{attr}})
               \gets\operatorname{ResponsibilityCredit}
               (x_i,a_i,e_i,G_i,M_t)$

        \If{$\kappa_i^{\mathrm{attr}}=1$}
            \State $M_t
                   \gets\operatorname{AccumulateFitness}
                   (M_t,R_i,\widehat{\mathbf{f}}_i,e_i)$
            \State $C_i^{\mathrm{fb}}
                   \gets\operatorname{GenerateOffspring}
                   (R_i,e_i,\mathcal{T}_i,G_i)$
        \ElsIf{$\kappa_i^{\mathrm{attr}}=2$}
            \State $C_i^{\mathrm{fb}}
                   \gets\operatorname{SpawnProvisional}
                   (e_i,\mathcal{T}_i,G_i)$
        \EndIf
    \EndIf
\EndIf

\State $C_t^{\mathrm{abs}}
       \gets\operatorname{PeriodicAbstract}(\mathcal{H}_t,M_t)$
\State $M_{t+1}
       \gets\operatorname{GovernAndSelect}_B
       (M_t\cup C_i^{\mathrm{fb}}\cup C_t^{\mathrm{abs}})$
\State \Return $M_{t+1}$

\end{algorithmic}
\end{algorithm}

\subsection{Responsibility Credit and Memory Evolution}
\label{sec:responsibility-credit}

The decision evidence $e_i$ identifies which content decisions in a generated material
are supported or contradicted by online outcomes.  A single generation may, however,
use multiple memories, so TIDE must identify which memories were responsible for those
decisions.  For each $m\in M_t$, responsibility requires that the memory was actually
used, is semantically linked to the target decision, and changes the corresponding
decision when removed:
\begin{equation}
\begin{aligned}
z_{i,m}^{\mathrm{resp}}
={}&\mathbb{I}\!\left[m\in\operatorname{Used}(G_i)\right]
   \mathbb{I}\!\left[\operatorname{Align}(m,e_i)\right] \\
 &\cdot
   \mathbb{I}\!\left[\operatorname{CF}
   (m;x_i,a_i,e_i,M_t)\right].
\end{aligned}
\label{eq:responsibility-credit}
\end{equation}

Here, $m$ denotes a memory entry, $\mathbb{I}[\cdot]$ is the indicator function,
$\operatorname{Used}(G_i)$ returns the memories actually used in trace $G_i$,
$\operatorname{Align}(m,e_i)$ tests whether $m$ is semantically linked to the content
decision described by $e_i$, and
$\operatorname{CF}(m;x_i,a_i,e_i,M_t)$ tests whether removing $m$ changes that
decision under counterfactual replay.  Thus, $z_{i,m}^{\mathrm{resp}}=1$ exactly when
all three conditions hold, and
$R_i=\{m\in M_t:z_{i,m}^{\mathrm{resp}}=1\}$ is the responsible set.  The set $R_i$
may contain multiple memories.  Memories that were retrieved but did not affect the
target decision receive no credit, whereas each memory that independently satisfies
the three responsibility conditions accumulates the corresponding contextual fitness
and supporting evidence.

The attribution state $\kappa_i^{\mathrm{attr}}\in\{0,1,2\}$ denotes unverifiable
responsibility, attribution to existing memories, and confirmed novelty, respectively.
We set $\kappa_i^{\mathrm{attr}}=1$ when at least one existing memory passes the
responsibility test.  We set $\kappa_i^{\mathrm{attr}}=2$ only when the decision
evidence and usage trace are valid but the decision is not covered by any existing
memory; this case may create a provisional candidate without a parent.  Otherwise,
$\kappa_i^{\mathrm{attr}}=0$, and no feedback-driven update is performed.

For attribution to existing memories, TIDE accumulates
$\widehat{\mathbf{f}}_i$ and $e_i$ as fitness evidence for each memory in $R_i$.
Conditional or contradictory evidence generates feedback-derived candidates
$C_i^{\mathrm{fb}}$ through narrowing, revision, or splitting.  Responsible memories
may be edited individually, while complementary memories may be combined through
crossover.  Confirmed novel evidence may instead produce a parentless provisional
candidate.  Periodic abstraction additionally generates $C_t^{\mathrm{abs}}$ from
valid recent records $\mathcal{H}_t$ and the current memory population $M_t$.  For
example, when an existing memory emphasizes price but repeated evidence supports
immediate usability only in a specific campaign context, TIDE creates a narrower
candidate rather than overwriting the original.

Existing memories and both candidate sources then undergo capacity-constrained
governance and selection:
\begin{equation}
M_{t+1}
=\mu_{\mathrm{TIDE}}(M_t,D_i,\mathcal{H}_t;B)
=\operatorname{GovernAndSelect}_B
\!\left(M_t\cup C_i^{\mathrm{fb}}\cup C_t^{\mathrm{abs}}\right).
\label{eq:tide-update}
\end{equation}
Here, $B$ is the maximum memory capacity, and
$\operatorname{GovernAndSelect}_B$ denotes validity filtering, deduplication,
conflict resolution, expiration, and capacity-constrained selection.  Each incumbent
memory retains its accumulated fitness evidence, while each candidate enters with
provisional fitness, provenance, and version information.  Subsequent feedback
determines its promotion, further narrowing, merging, or eviction.  This enables
memory evolution through outcome evidence and cross-memory abstraction without
unbounded accumulation.

\section{Experiments}
\label{sec:experiments}
We evaluate TIDE in real-world e-commerce content generation and on a
public delayed-label benchmark.  Our experiments address four research
questions spanning end-to-end effectiveness, feedback interpretation,
responsibility attribution, and cross-task generalization.  Detailed
experimental settings are provided in the appendix.

\begin{itemize}[leftmargin=*, labelsep=0.5em, topsep=3pt, itemsep=2pt,
                 parsep=0pt, partopsep=0pt]
    \item \textbf{(RQ1)} Does feedback-evolved memory improve offline
    generation and online performance?

    \item \textbf{(RQ2)} How can delayed, context-dependent outcomes become
    actionable generation feedback?

    \item \textbf{(RQ3)} How can feedback be attributed to responsible
    memories and translated into targeted edits?

    \item \textbf{(RQ4)} Do feedback-driven memory updates generalize to
    public delayed-label tasks?
\end{itemize}

\subsection{Main Results (RQ1)}
\label{sec:rq1}

\paragraph{Offline results.}

To answer the offline part of RQ1, we examine whether memory evolved from historical delivery feedback improves generation on strictly future requests. Within each backbone, all methods use the same data, request order, feedback budget, and candidate-set size, and all outputs are evaluated by \texttt{DeepSeek-V4-Pro}. We compare TIDE with representative external-memory methods on \texttt{qwen3.5-27b} and \texttt{Qwen3-4B}. On \texttt{Qwen3-4B}, we additionally include SFT, GRPO~\citep{shao2024deepseekmath}, and MMPO~\citep{mao2026betterspurstartobjective} as parameter-based self-evolution baselines. We evaluate generation quality, paired utility gains, negative transfer, adaptation compute, and inference overhead.
As shown in Table~\ref{tab:offline_main}, TIDE consistently provides the best overall performance among the memory methods. On \texttt{qwen3.5-27b}, it achieves the highest compliance, format, efficiency, diversity, and joint-pass scores, with a MEG of $+7.75$ and the lowest inference overhead of 750 tokens per request, while requiring adaptation compute comparable to other memory methods. On \texttt{Qwen3-4B}, TIDE remains the strongest memory method, although all memory-only approaches are limited by the backbone's weak format-following capability. Parameter-based post-training alleviates this limitation but incurs adaptation costs several orders of magnitude higher. Combining TIDE with SFT further improves overall generation quality and yields a MEG of $+6.71$ with only 0.118 additional PFLOPs beyond SFT. These results answer the offline part of RQ1 affirmatively: feedback-evolved memory improves future generation with capable backbones, while complementing parameter-based training at substantially lower incremental adaptation cost.

\begin{table*}[tbp]
\centering
\caption{Offline generation quality and adaptation efficiency.}
\label{tab:offline_main}
\fontsize{8}{10}\selectfont
\setlength{\tabcolsep}{2.5pt}
\renewcommand{\arraystretch}{1.1}
\begin{tabular*}{\textwidth}{@{\extracolsep{\fill}}lrrrrrrrrr@{}}
\toprule
\textbf{Method}
& \multicolumn{5}{c}{\textbf{Generation quality}}
& \multicolumn{2}{c}{\textbf{Memory evolution}}
& \multicolumn{2}{c}{\textbf{Efficiency}} \\
\cmidrule(lr){2-6}\cmidrule(lr){7-8}\cmidrule(l){9-10}
& \shortstack{Comp. $\uparrow$\\(\%)}
& \shortstack{Fmt. $\uparrow$\\(\%)}
& \shortstack{Eff. $\uparrow$\\(\%)}
& \shortstack{Div. $\uparrow$\\(0--100)}
& \shortstack{Joint $\uparrow$\\(\%)}
& MEG $\uparrow$
& \shortstack{Neg. Trans. $\downarrow$\\(\%)}
& \shortstack{Adapt. $\downarrow$\\(PFLOPs)}
& \shortstack{Inference $\downarrow$\\(tokens/req.)} \\
\midrule
\multicolumn{10}{@{}l}{\textbf{Backbone:} \texttt{qwen3.5-27b}} \\[2pt]
No Memory
& 56.0 & 79.2 & 56.0 & 65.5 & 12.0
& 0.00 & 0.0 & -- & -- \\
Initial Memory (Frozen)
& 62.0 & 83.6 & 80.0 & 61.5 & 20.3
& -1.88 & 22.0 & 0.803 & 1,271 \\
Full History
& 57.0 & 77.0 & 46.0 & 68.0 & 12.0
& -1.13 & 30.0 & 0.973 & 1,968 \\
ExpRAG
& 58.0 & 77.0 & 32.0 & 59.0 & 12.0
& -2.50 & 32.0 & 0.876 & 1,967 \\
Reflexion
& 61.0 & 79.2 & 46.0 & 69.0 & 14.8
& +1.25 & 18.0 & 0.768 & 828 \\
AWM
& 53.0 & 77.0 & 54.0 & 65.0 & 12.0
& -0.75 & 30.0 & 0.778 & 918 \\
Mem0
& 57.0 & 81.4 & 54.0 & 64.5 & 12.0
& +0.88 & 24.0 & 0.779 & 830 \\
A-MEM
& 56.0 & 77.0 & 44.0 & 65.5 & 12.0
& -2.88 & 34.0 & 0.825 & 1,696 \\
\textbf{TIDE}
& \textbf{63.0}
& \textbf{88.0}
& \textbf{84.0}
& \textbf{69.5}
& \textbf{23.0}
& \textbf{+7.75}
& \textbf{14.0}
& 0.774
& \textbf{750} \\
\midrule
\multicolumn{10}{@{}l}{\textbf{Backbone:} \texttt{Qwen3-4B}} \\[2pt]
No Memory
& 28.0 & 0.0 & 4.0 & 49.5 & 0.0
& 0.00 & 0.0 & -- & -- \\
Initial Memory (Frozen)
& 31.0 & 0.0 & 16.0 & 48.0 & 0.0
& -16.00 & 78.0 & 0.121 & 1,494 \\
Full History
& 32.0 & 0.0 & 22.0 & 47.5 & 0.0
& -0.63 & 30.0 & 0.149 & 2,000 \\
ExpRAG
& 33.0 & 0.0 & 22.0 & \textbf{53.5} & 0.0
& +1.50 & 58.0 & 0.132 & 2,000 \\
Reflexion
& 28.0 & 0.0 & 20.0 & 26.5 & 0.0
& -5.88 & 60.0 & 0.115 & 1,176 \\
AWM
& 30.0 & 0.0 & 18.0 & 43.0 & 0.0
& +0.50 & 26.0 & 0.117 & 1,241 \\
Mem0
& 33.0 & 0.0 & 22.0 & 44.0 & 0.0
& -0.75 & 32.0 & 0.116 & 1,215 \\
A-MEM
& 31.0 & 0.0 & 16.0 & 47.0 & 0.0
& -1.13 & 30.0 & 0.125 & 1,905 \\
\textbf{TIDE}
& \textbf{35.0}
& \textbf{0.0}
& \textbf{26.0}
& \textbf{53.5}
& \textbf{0.0}
& \textbf{+1.75}
& \textbf{24.0}
& 0.118
& \textbf{1,046} \\
\cmidrule(l){1-10}

SFT
& 50.0 & 66.0 & 42.0 & 56.0 & 16.0
& +4.63 & 8.0 & 153 & 0 \\
GRPO
& 49.0 & 68.0 & 20.0 & 56.0 & 8.0
& +2.25 & 10.0 & 1,433 & 0 \\
MMPO
& 52.0 & 71.0 & 30.0 & 63.0 & 8.0
& +2.63 & 8.0 & 1,421 & 0 \\
\textbf{SFT + TIDE}
& \textbf{56.0}
& \textbf{72.5}
& \textbf{43.8}
& \textbf{65.9}
& \textbf{18.0}
& \textbf{+6.71}
& \textbf{6.7}
& 153.118
& 783 \\
\bottomrule
\end{tabular*}
\end{table*}

\paragraph{Online results.}

\begin{table}[t]

\caption{
Relative uplift in the one-week online comparison.
}
\label{tab:ab-main}

\centering
\small
\setlength{\tabcolsep}{4pt}
\begin{tabular}{lr}
\toprule
Metric & TIDE Rel. Uplift \\
\midrule
UCTR            & \textbf{+5.10\%} \\
Activation Rate & \textbf{+4.79\%} \\
\bottomrule
\end{tabular}

\end{table}

We further conduct a one-week randomized online A/B experiment to
examine whether TIDE delivers measurable gains in real-world deployment. Under comparable traffic conditions, we compare a
baseline material pool with an experimental pool augmented with
TIDE-evolved materials, while keeping the product pool, placement,
and delivery mechanism fixed. For any rate metric $r$, relative
uplift is defined as
$(r_{\mathrm{TIDE}}/r_{\mathrm{Baseline}}-1)\times100\%$.
As shown in Table~\ref{tab:ab-main}, the experimental pool yields
statistically significant relative uplifts of 5.10\% in UCTR and
4.79\% in activation rate (both $p<0.001$). These results provide
empirical evidence that deploying TIDE-evolved materials is
associated with improved click and activation outcomes under
comparable real-world traffic.

\subsection{Aligning Online Outcomes with Feedback Memory (RQ2)}
\label{sec:rq2}

We chronologically replay deployment logs and make
outcomes available only after they have matured, thereby preventing
future-information leakage. Historical cases are aligned by channel,
benefit, and semantic context, and prediction--outcome discrepancies
are distilled into actionable generation feedback. Across 28
channel--benefit tasks, we compare an LLM rubric, direct LLM
prediction, BERT, a distilled recommender, and TIDE. Pre-launch
prediction is evaluated using macro-averaged rank correlation and
pairwise accuracy, while downstream utility is measured against the
no-memory condition using Future MEG, win rate, and negative-transfer
rate under matched generation settings.

\begin{figure*}[tbp]
    \centering
    \includegraphics[width=\textwidth]
    {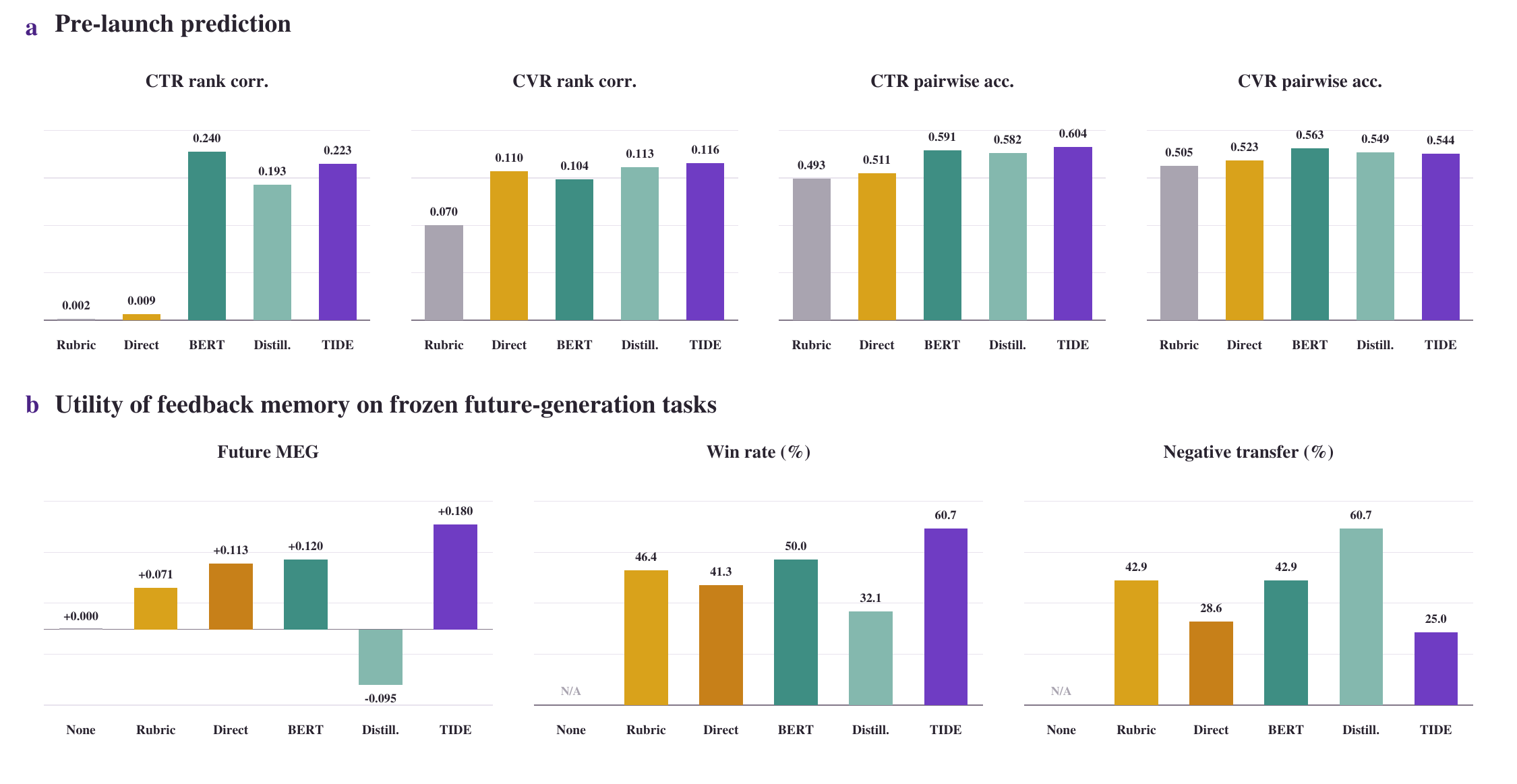}

    \caption{
    Pre-launch CTR/CVR ranking and downstream feedback-memory utility.
    }
    \label{fig:rq2_temporal_semantic_alignment}
\end{figure*}

As shown in Figure~\ref{fig:rq2_temporal_semantic_alignment}, TIDE
achieves the highest CVR rank correlation ($0.116$) and CTR pairwise
accuracy ($0.604$), while BERT leads the other two ranking metrics but
requires parameter adaptation. More importantly, TIDE obtains the
largest observed Future MEG ($+0.180$), the highest win rate
($60.7\%$), and the lowest negative-transfer rate ($25.0\%$), yielding
the strongest overall downstream utility profile. Taken together, these results indicate that TIDE effectively converts delayed online outcomes into actionable feedback for improving future generation, thereby answering RQ2.

\subsection{From Aligned Feedback to Memory Evolution (RQ3)}
\label{sec:rq3}

\begin{figure}[t]

    \centering
    \includegraphics[width=\linewidth]{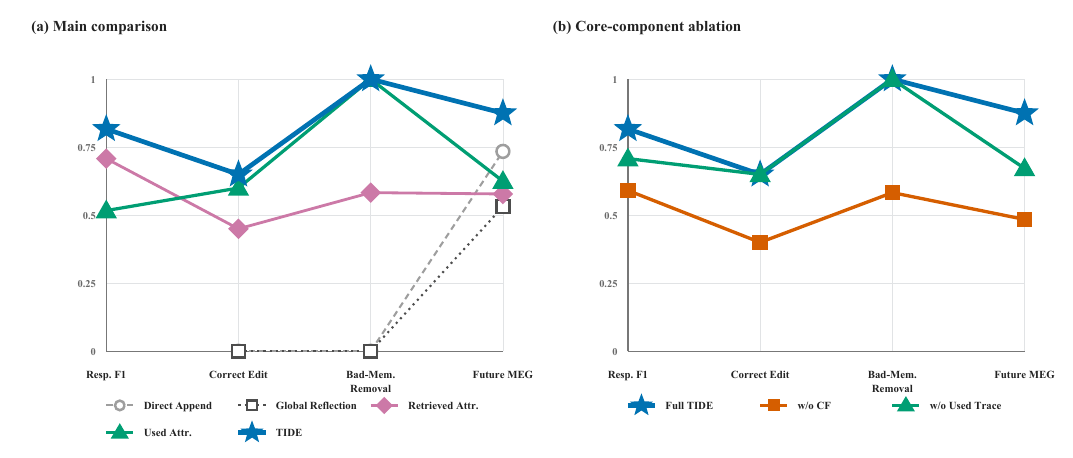}

    \caption{
    Responsibility attribution and directed memory editing, with all
    metrics direction-adjusted so that higher values indicate better
    performance.
    }
    \label{fig:rq3_memory_evolution}

\end{figure}

We evaluate whether aligned feedback can identify responsible memories
and guide targeted edits that improve future generation. Across 20
controlled episodes with five injected noise types, all methods use the
same feedback, initial memory, and task order. TIDE combines actual
memory-usage traces with deterministic leave-one-out counterfactual
replay and edits a memory only when its removal improves utility.

As shown in Figure~\ref{fig:rq3_memory_evolution}(a), TIDE achieves the
strongest overall performance, with a Responsibility F1 of $0.817$, a
Correct Edit rate of $0.650$, zero Bad-Memory Survival, and a Future
MEG of $+0.0750$. Compared with Retrieved-Memory Attribution, TIDE
improves responsibility localization, edit accuracy, and future utility
while eliminating bad-memory survival.

The ablations in Figure~\ref{fig:rq3_memory_evolution}(b) confirm the
contributions of both components. Removing counterfactual replay causes
the largest degradation, reducing Responsibility F1 and Correct Edit
to $0.591$ and $0.400$, increasing Bad-Memory Survival to $0.417$, and
lowering Future MEG to $-0.0031$. Removing usage traces reduces
Responsibility F1 to $0.708$ and Future MEG to $+0.0344$. Taken
together, these results answer RQ3 by showing that usage traces narrow
responsibility localization, while counterfactual replay validates
responsibility and enables effective memory edits for future
generation.

\subsection{Generalization to Delayed-Label Tasks (RQ4)}
\label{sec:rq4}

We evaluate whether TIDE generalizes beyond its original generation
domain to delayed-feedback tasks. To represent delayed and progressively
available outcomes, we adapt MemoryCD into a chronological interaction
stream and expose each label only after its prescribed delay. This
protocol ensures that each method updates its memory using only feedback
available at the current time. We compare TIDE with static memory,
full-history prompting, retrieval- and reflection-based agents, and
recent memory-management methods. Prediction quality is evaluated using
MAE and RMSE, memory utility using MEG@$\delta{=}5$, delay robustness
using $\Delta_{\mathrm{delay}}$, and reliability using user-level
negative transfer. Memory size and LLM tokens per feedback-equivalent
update are also reported.

As shown in Table~\ref{tab:memorycd}, TIDE achieves the lowest MAE
($0.531$) and RMSE ($0.895$), outperforming the strongest non-TIDE
results of $0.572$ and $1.073$, respectively, and obtains the highest
MEG@$\delta{=}5$ ($+0.0365$). Among adaptive methods, TIDE is also the
least sensitive to delayed labels, with
$\Delta_{\mathrm{delay}}=+0.0019$, and ties Mem0 for the lowest
user-level negative-transfer rate ($16.7\%$). TIDE uses 48,603 memory
tokens, fewer than the approximately 50--55K tokens used by the other
adaptive memory methods. Its 2,693.3 LLM tokens per feedback-equivalent
update remain substantially below FullHistory while staying comparable
to the retrieval-based methods. It also maintains controlled adaptation
overhead. Taken together, these results answer RQ4 affirmatively: TIDE
generalizes beyond the original generation setting and provides the
strongest overall profile in prediction accuracy, memory utility, delay
robustness, and user-level reliability on the delayed-label task.

\begin{table*}[t]
\centering
\caption{Results on the MemoryCD delayed-label benchmark.}
\label{tab:memorycd}

\small
\renewcommand{\arraystretch}{1.18}
\setlength{\tabcolsep}{4.5pt}

\resizebox{\textwidth}{!}{%
\begin{tabular}{@{}lrrrrrrr@{}}
\toprule
\textbf{Method}
& \multicolumn{1}{c}{\textbf{MAE} $\downarrow$}
& \multicolumn{1}{c}{\textbf{RMSE} $\downarrow$}
& \multicolumn{1}{c}{\shortstack{\textbf{MEG@$\delta{=}5$}\\$\uparrow$}}
& \multicolumn{1}{c}{\shortstack{
    $\boldsymbol{\Delta_{\mathrm{delay}}}$\\
    $(|\cdot|\downarrow)$
}}
& \multicolumn{1}{c}{\shortstack{
    \textbf{User Neg.}\\
    \textbf{Trans.} $\downarrow$
}}
& \multicolumn{1}{c}{\shortstack{
    \textbf{Memory}\\
    \textbf{Tok.}
}}
& \multicolumn{1}{c}{\shortstack{
    \textbf{LLM Tok.}/\\
    \textbf{Feedback-eq.} $\downarrow$
}} \\
\midrule

StaticMemory ($\mu_0$)
& 0.676 & 1.204 & 0.0000 & +0.0000
& 0.0\% & 0 & 838.3 \\

FullHistory$\dagger$
& 0.575 & 1.073 & +0.0253 & +0.0163
& 33.3\% & 54,875 & 15,273.2 \\

RAG \citep{lewis2020rag}
& 0.587 & 1.131 & +0.0222 & +0.0205
& 25.0\% & 54,875 & 2,438.2 \\

ReAct \citep{yao2023react}
& 0.610 & 1.152 & +0.0167 & +0.0035
& 25.0\% & 51,022 & 2,438.1 \\

Reflexion \citep{shinn2023reflexion}
& 0.606 & 1.156 & +0.0177 & +0.0153
& 25.0\% & 54,875 & 2,438.1 \\

Self-RAG \citep{asai2024selfrag}
& 0.653 & 1.180 & +0.0059 & -0.0066
& 33.3\% & 51,254 & 2,427.6 \\

Mem0 \citep{chhikara2025mem0}
& 0.572 & 1.103 & +0.0260 & +0.0215
& \textbf{16.7\%} & 51,026 & 2,425.7 \\

A-MEM \citep{xu2025amem}
& 0.611 & 1.149 & +0.0163 & +0.0162
& 25.0\% & 51,440 & 2,421.5 \\

AWM \citep{wang2025awm}
& 0.619 & 1.148 & +0.0142 & +0.0042
& 33.3\% & 50,882 & 2,427.7 \\

DC-Cu \citep{suzgun2026dynamic}
& 0.682 & 1.177 & -0.0014 & +0.0146
& 33.3\% & 50,129 & \textbf{838.0} \\

ExpRAG \citep{wei2025evomemory}
& 0.615 & 1.179 & +0.0153 & +0.0049
& 25.0\% & 51,407 & 2,437.2 \\

\addlinespace[2pt]
\textbf{TIDE (Ours)}
& \textbf{0.531}
& \textbf{0.895}
& \textbf{+0.0365}
& \textbf{+0.0019}
& \textbf{16.7\%}
& 48,603
& 2,693.3 \\
\bottomrule
\end{tabular}%
}
\end{table*}

\section{Conclusion}

We study how delayed, context-dependent recommendation outcomes can be
converted into memory for future tasks. We introduce MEG to measure
genuine adaptation and propose TIDE, combining temporal, semantic, and
responsibility credit to evolve a capacity-constrained external memory
with a frozen generator. TIDE achieves the strongest overall
performance among evaluated memory methods, including a $+7.75$
offline MEG, and its evolved materials are associated with online gains
of $5.10\%$ in UCTR and $4.79\%$ in activation rate. Further analyses
validate feedback alignment and responsible editing, while MemoryCD
results demonstrate generalization to delayed-label tasks. TIDE
provides an effective, efficient, and traceable approach to continual
agent adaptation through feedback-driven memory evolution under delayed
and context-dependent outcome feedback.

\FloatBarrier

\section*{AI Use Disclosure}

Generative AI tools were used to improve the clarity, grammar, and
readability of the manuscript, as well as to assist with software
development, including code completion, debugging, and refactoring.
All AI-assisted text was reviewed and revised by the authors. All
AI-assisted code was manually inspected and validated through testing
and experimental verification. The authors take full responsibility
for the correctness, originality, claims, results, and final content
of this work.

\bibliography{references}
\bibliographystyle{iclr2027_conference}

\clearpage
\appendix
\renewcommand{\theHsection}{appendix.\Alph{section}}
\FloatBarrier

\FloatBarrier
\section{Additional Experimental Configuration}
\suppressfloats[t]
\label{app:experimental_configuration}

\noindent\textbf{Unified protocol and main offline experiment.}
The 88VIP experiments are partitioned chronologically into memory initialization, feedback evolution, validation, and strictly future testing. At time $t$, a method may access only feedback that has already arrived, and any sample used to construct memory or train model parameters is excluded from future testing. Methods using the same backbone share the task order, feedback budget, generation constraints, candidate-set size, and test samples; memory-based methods additionally share the initial memory, retriever, retrieval count, and capacity limit. With \texttt{qwen3.5-27b}, we compare No Memory, Initial Memory (Frozen), Full History, ExpRAG, Reflexion, AWM, Mem0, A-MEM, and TIDE. The \texttt{Qwen3-4B} experiments additionally include SFT, GRPO, MMPO, and SFT + TIDE. All generated outputs are evaluated by a frozen \texttt{DeepSeek-V4-Pro} judge. Method identifiers are removed from the evaluation inputs, and all methods use the same evaluation prompt and scoring criteria. The judge evaluates Compliance, Format, Efficiency, and Diversity. Each dimension is normalized to a 0--100 scale, and task utility is their equally weighted mean. The main experiment contains 50 frozen future requests, with ten materials generated for each request. MEG is computed as the paired request-level utility difference relative to No Memory, and Negative Transfer is the proportion of requests for which a method obtains lower utility than No Memory. Thus, No Memory has a MEG of zero by definition and is shown as the reference condition. Statistical comparisons preserve the request-level pairing among methods. For methods using the same backbone, we convert training and inference costs into estimated FLOPs under a unified dense-Transformer accounting convention: a forward pass over $T$ tokens is approximated as $2PT$, and full-parameter training as $6PT$, where $P$ is the number of model parameters. These values are approximate accounting estimates rather than measurements of wall-clock time or hardware efficiency. Parameter updates, inference calls, and memory-construction calls using the same model can therefore be compared under a consistent accounting convention. For heterogeneous systems such as BERT, conventional recommender models, and LLMs, whose architectures, operator compositions, and workloads differ, we do not perform uncalibrated cross-system rankings using FLOPs, hardware resources, or token counts. Instead, we report estimates under their respective accounting conventions and restrict direct compute comparisons to methods using the same model and computation rule.

\noindent\textbf{Online A/B experiment.}
The randomized online A/B experiment ran for one week. Eligible traffic was randomly assigned through the platform's prespecified traffic-allocation mechanism. The control group used the existing online material pool supported by directly accumulated experience memory, whereas the treatment group introduced TIDE-generated materials while keeping the base model, product pool, eligible audience, placement, number of candidates, and delivery mechanism fixed. Before outcome analysis, three prespecified experimental buckets with nonzero exposure to TIDE-generated materials were designated as treatment buckets and pooled to estimate the overall treatment effect. They were compared with a control bucket containing no TIDE-generated materials. The AA bucket was used only to verify the stability of the traffic-allocation mechanism and was not included in the treatment group or the treatment-effect estimate. No experimental group was selected, removed, or reassigned retrospectively based on observed outcomes. We report relative lifts in UCTR and Activation Rate, defined for a rate metric $r$ as $(r_{\mathrm{Treatment}}/r_{\mathrm{Control}}-1)\times100\%$. Approximate 95\% confidence intervals and two-sided significance tests are computed from the corresponding group counts and rates using the delta method for relative risks. Absolute traffic volumes and business rates are retained for internal audit because of confidentiality requirements; only relative effects are reported in the paper.

\noindent\textbf{Online--offline alignment, predictive baselines, and feedback memory.}
The RQ2 ranking experiment contains 28 equally weighted channel--benefit tasks, with five candidate materials per task and 140 materials in total. LLM Rubric, Direct LLM, Lifecycle, and TIDE all use a frozen \texttt{qwen3.5-27b}. Lifecycle retrieves relevant cases exclusively from historical materials whose outcomes have matured before the prediction time, whereas TIDE uses the same temporally available historical evidence to construct a two-layer feedback memory consisting of abstract experience cards and traceable source cases. BERT and the distilled recommender represent two non-generative baselines based on content-semantic modeling and exposure-feedback modeling, respectively. BERT is trained only on historical online materials available before the prediction cutoff and is evaluated on the same frozen future candidate sets as the other methods. Channel, benefit, audience, style, and copy text are serialized in a fixed order; channel names retain their placement labels, other multi-valued attributes are normalized and joined with ``/'', and missing attributes and channels are represented by ``none'' and ``unknown'', respectively. This representation jointly encodes material semantics and delivery context. The distilled recommender is likewise trained only on matured exposure feedback available before the prediction cutoff. It jointly embeds discrete and continuous features and predicts the CTR and CVR of each candidate material. All training examples, feature construction, memory construction, and label aggregation strictly obey the temporal cutoff, preventing information observed after the prediction time from entering model construction. Ranking performance is macro-averaged over tasks so that each channel--benefit task receives equal weight. All 28 tasks are retained for CTR ranking, whereas the CVR evaluation excludes five tasks whose observed CVR values are identical for all candidates within the task and therefore induce no valid ranking. In the feedback-memory experiment, we fix the generator, prompt body, decoding parameters, number of candidates, and 6,000-character context limit, varying only the feedback memory. Each condition generates five materials per task. Method identities are removed before evaluation, and outputs are evaluated under the same presentation and scoring protocol. Task utility is the equally weighted mean of constraint satisfaction, clarity, benefit accuracy, and persuasiveness, each scored from 0 to 10. Future MEG, win rate, and Negative Transfer are computed relative to the matched No Memory condition. Uncertainty is estimated using 10,000 paired task-level bootstrap samples, with all outputs belonging to the same task resampled together.

\noindent\textbf{Responsibility attribution and memory editing.}
RQ3 uses 20 controlled mock requests and injects five types of noise into the initial candidate memory pool: near duplicates, stale memories, conflicting memories, cross-scenario hard negatives, and misleading feedback. The complete candidate pool contains approximately 6,009--6,151 estimated tokens, while the injected-noise portion is constrained to at most 2,000 tokens. The injected memories and their intended editing operations provide controlled ground truth for evaluating responsibility localization and memory editing. Responsibility F1 measures whether a method identifies the ground-truth responsible memories, while Correct Edit requires both the memory identity and the corresponding editing operation to be correct. Bad-Memory Survival is computed on the 12 requests involving stale, conflicting, or misleading memories, for which removal or revision of an explicitly harmful memory is defined by construction. Future MEG is computed by pairing each method with the No Memory condition on the same frozen future requests. Each method generates ten materials for each of the 20 requests. Method identifiers are removed from the judge input, and all outputs are evaluated by the same frozen \texttt{DeepSeek-V4-Pro} judge using a common rubric. Uncertainty is estimated using 10,000 paired request-level bootstrap samples, with the request serving as the resampling unit. Counterfactual responsibility analysis performs deterministic leave-one-memory-out replay with temperature 0. It separately records whether a memory was retrieved, whether it was referenced during generation, and whether its removal changes the generated output or improves its evaluated utility. An edit is triggered only when the leave-one-memory-out intervention produces a positive utility change, thereby distinguishing mere retrieval or reference from intervention-supported responsibility.

\noindent\textbf{Public delayed-label benchmark.}
For the public evaluation, we adapt MemoryCD~\citep{zhang2026memorycd} into a chronological delayed-label benchmark while retaining its original samples, labels, user histories, and temporal order. We control only the timing of label release, memory updates, and future evaluation; therefore, the experiment introduces a delayed-label evaluation protocol rather than claiming that MemoryCD contains natively delayed labels. A ground-truth label is released $\delta$ task steps after its prediction, with $\delta=5$ used as the main setting. For each user, the first 80\% of chronologically ordered interactions support memory evolution. A cooldown interval of length $\delta$ then releases feedback that remains pending from this evolution period. After the cooldown, memory is frozen and evaluated on the final 20\% of interactions; labels from this final test segment are never used to update memory. We report MAE, RMSE, and paired MEG under $U_i=1-|\hat{y}_i-y_i|/4$, where the denominator corresponds to the maximum absolute error on the five-point rating scale. In accordance with the main RQ4 evaluation, StaticMemory is the reference policy for MEG on MemoryCD and therefore has a MEG of zero by definition. We additionally report sensitivity to feedback delay, user-level Negative Transfer, frozen-memory tokens, and LLM tokens per feedback-equivalent update. All methods share the chronological task order, available-feedback budget, retrieval count, and memory-capacity constraint. Statistical uncertainty is estimated using a user-cluster bootstrap so that all interactions belonging to the same user are resampled together, preserving within-user dependence.

\FloatBarrier
\section{Multi-Round Memory Evolution and Convergence}
\suppressfloats[t]
\label{app:e1_multiround_evolution}

\paragraph{Experimental setup.}
E1 examines whether TIDE can continuously improve generation quality as new
feedback arrives while avoiding the unbounded memory growth of Full History
and Append-only. We compare Full History, Append-only, Random Replacement,
and TIDE over Rounds 1--10 on 12 stratified tasks with seed 13. For every
method--task--round combination, we generate ten materials and score them
jointly with a frozen judge. The experiment comprises 480 candidate sets,
4,800 materials, and 480 successful judge calls. Each batch is further divided
into five A/B units matched by task and slot, and we record
\texttt{assigned\_memory\_ids} to compute paired responsibility credit. We report Mean Utility, defined as the mean of the normalized multidimensional
scores across individual materials, together with Joint Pass, the four
component scores, Memory Tokens, paired A--B effects, and generation and judge
token consumption.

\begin{figure*}[htbp]
  \centering
  \includegraphics[width=0.82\textwidth]{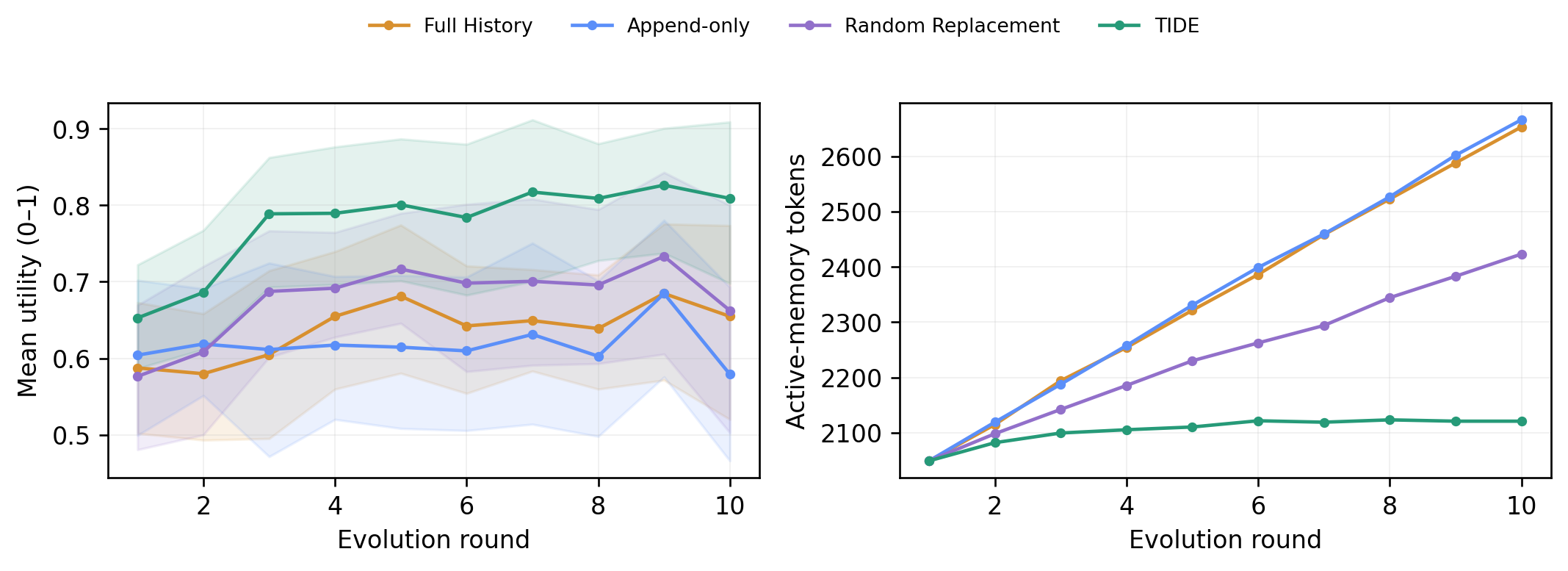}

  \caption{Ten-round evolution trajectories on 12 fixed tasks with seed 13.
  Left: mean normalized utility with task-bootstrap 95\% confidence intervals.
  Right: active-memory tokens. TIDE maintains a bounded 80-item population,
  whereas Full History and Append-only grow throughout replay.}
  \label{fig:e1_evolution_trajectory}
\end{figure*}

\paragraph{Results.}
Figure~\ref{fig:e1_evolution_trajectory} shows the ten-round quality and memory
trajectories. TIDE achieves the strongest final utility while maintaining a
bounded memory population, in contrast to the continual growth of Full
History and Append-only.

\begin{figure*}[htbp]
  \centering
  \includegraphics[width=0.82\textwidth]{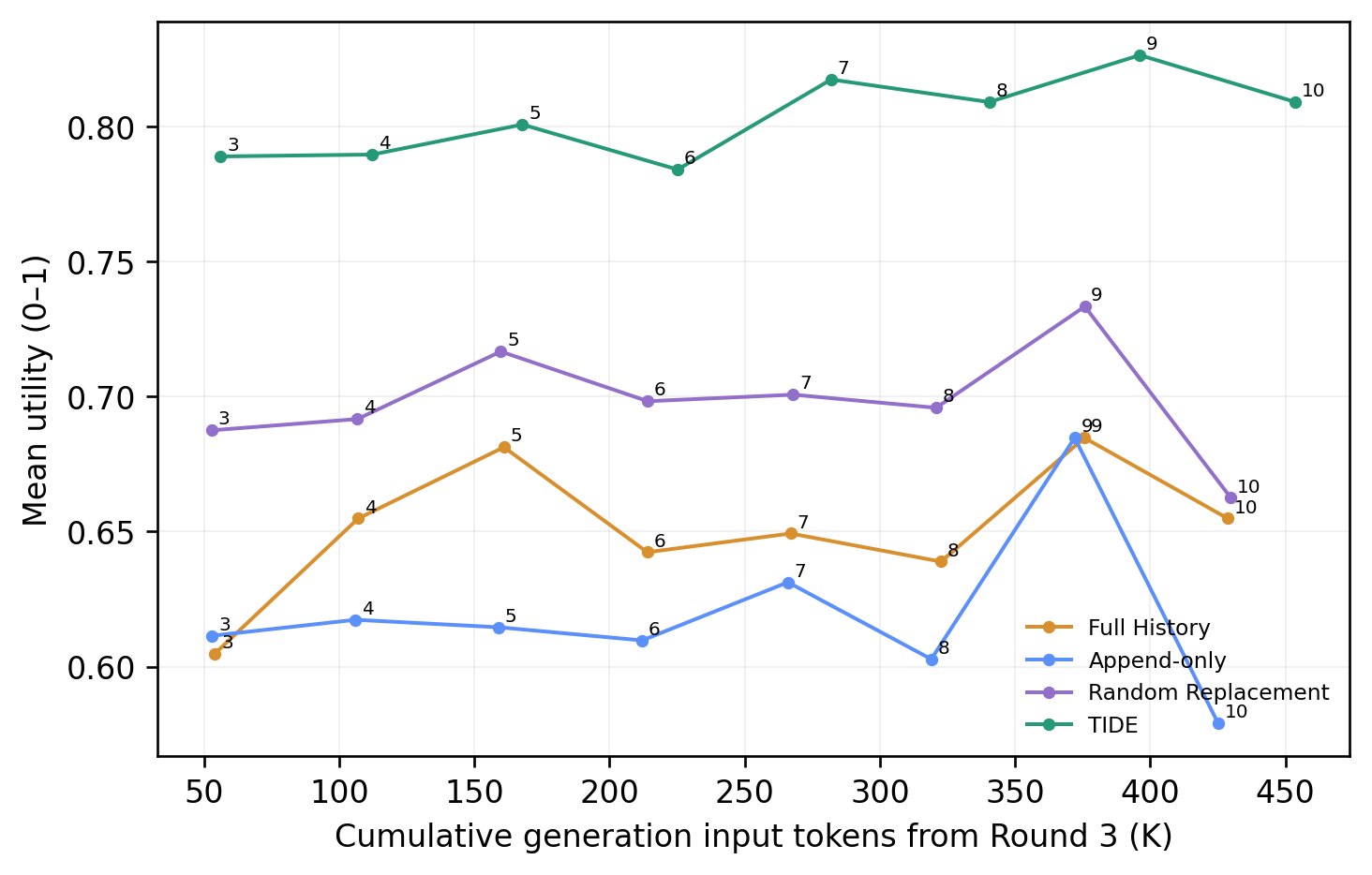}

  \caption{Diagnostic quality--cost trajectories from Rounds 3--10. The
  horizontal axis reports exact cumulative generation input tokens, and the
  vertical axis reports mean normalized utility. Numbers indicate replay
  rounds. Rounds 1--2 are omitted because per-job generation-token telemetry
  was not recorded during the original run.}
  \label{fig:e1_quality_cost}
\end{figure*}

\begin{figure*}[htbp]
  \centering
  \includegraphics[width=0.82\textwidth]{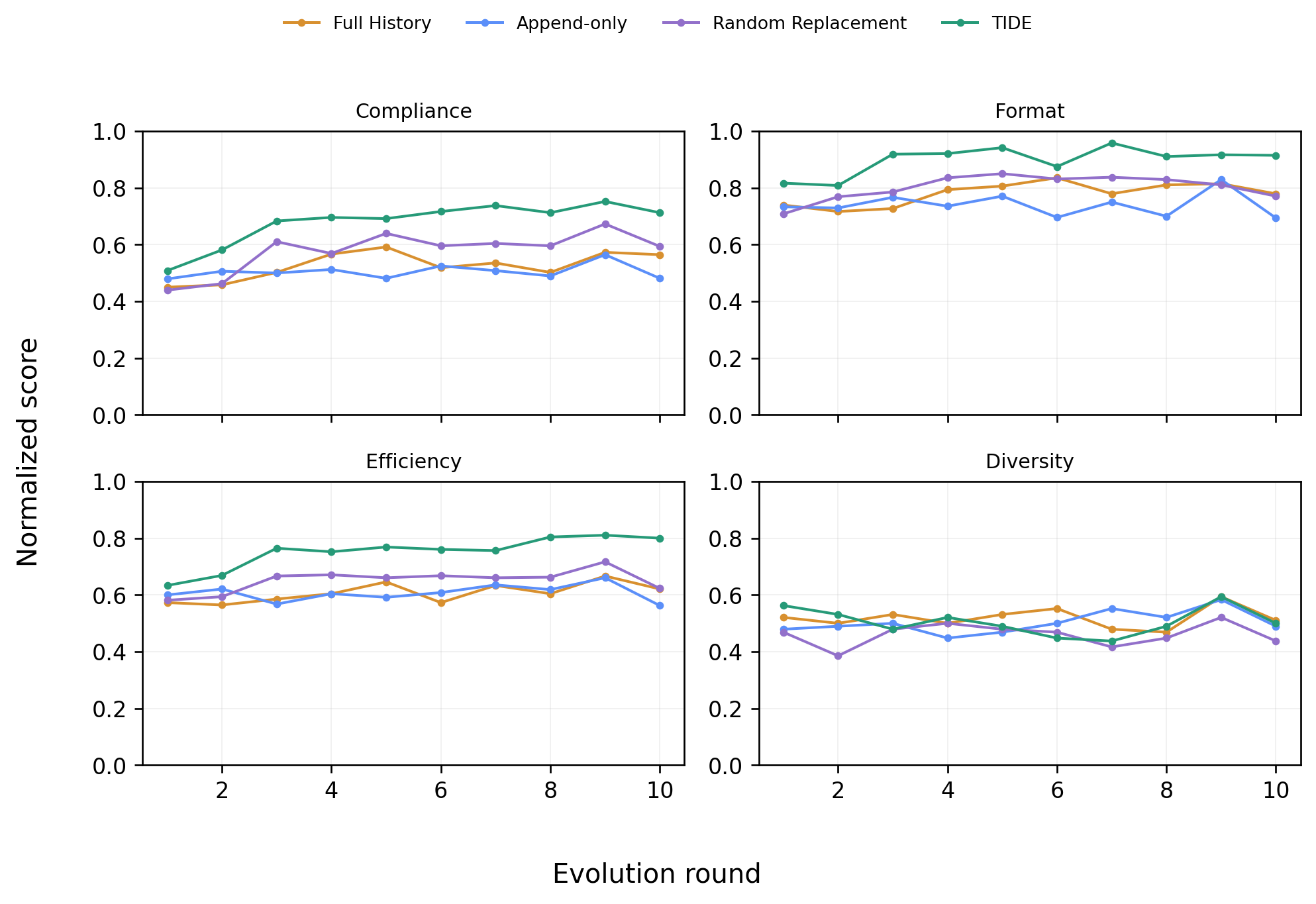}

  \caption{Normalized Compliance, Format, Efficiency, and batch-level
  Diversity over ten replay rounds. TIDE's advantage is concentrated in the
  first three dimensions; its final Diversity is comparable to that of the
  other methods rather than uniformly higher.}
  \label{fig:e1_subdimension_trajectory}
\end{figure*}

\begin{figure}[htbp]
  \centering
  \includegraphics[width=0.60\textwidth]{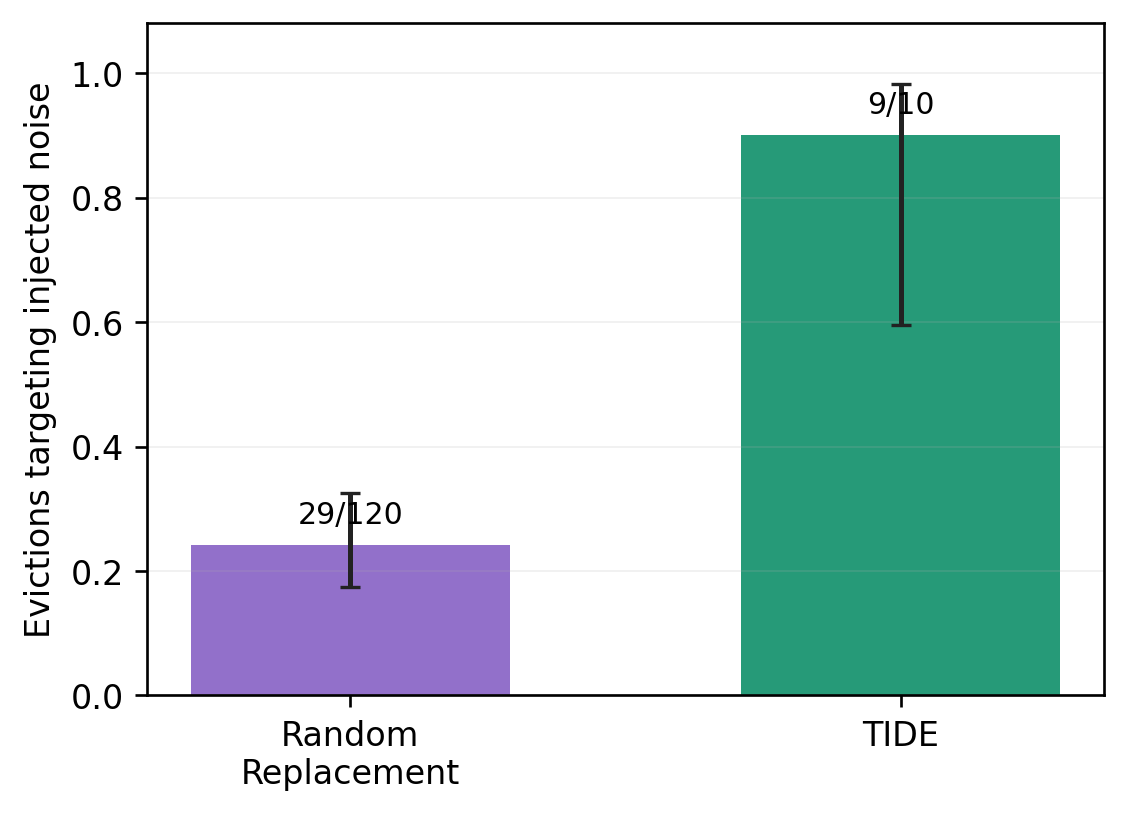}
  \caption{Selectivity of eviction operations over the ten-round E1 replay.
  Bars report the proportion of all evictions that target injected noisy
  memories; error bars denote Wilson 95\% confidence intervals. TIDE targets
  injected noise in 9 of 10 evictions, compared with 29 of 120 for Random
  Replacement.}
  \label{fig:e1_eviction_selectivity}
\end{figure}

\begin{table*}[htbp]
\centering
\caption{
Final-round generation quality and memory cost after ten feedback
rounds. Mean Utility averages normalized compliance, format, and
efficiency scores; 95\% CIs are computed by task-level bootstrap over
12 fixed tasks.
}
\label{tab:e1_final_results}

\small
\setlength{\tabcolsep}{4pt}
\renewcommand{\arraystretch}{1.12}

\resizebox{\textwidth}{!}{%
\begin{tabular}{@{}lrcrrrrrr@{}}
\toprule
\textbf{Method}
& \shortstack{\textbf{Mean}\\\textbf{Utility} $\uparrow$}
& \textbf{95\% CI}
& \textbf{Comp.} $\uparrow$
& \textbf{Fmt.} $\uparrow$
& \textbf{Eff.} $\uparrow$
& \textbf{Div.} $\uparrow$
& \shortstack{\textbf{Joint}\\\textbf{Pass} $\uparrow$}
& \shortstack{\textbf{Memory}\\\textbf{Tok.} $\downarrow$} \\
\midrule

Full History
& 0.655
& [0.520, 0.774]
& 0.565
& 0.779
& 0.621
& 0.510
& 0.442
& 2,653 \\

Append-only
& 0.579
& [0.466, 0.693]
& 0.481
& 0.694
& 0.562
& 0.490
& 0.367
& 2,666 \\

Random Replacement
& 0.663
& [0.503, 0.800]
& 0.594
& 0.771
& 0.623
& 0.438
& 0.500
& 2,423 \\

\addlinespace[2pt]
\textbf{TIDE}
& \textbf{0.809}
& \textbf{[0.698, 0.909]}
& \textbf{0.713}
& \textbf{0.915}
& \textbf{0.800}
& 0.500
& \textbf{0.617}
& \textbf{2,121} \\
\bottomrule
\end{tabular}%
}
\end{table*}

After ten feedback rounds, TIDE demonstrates consistent advantages in
generation quality, memory efficiency, and selective editing. First, it
achieves the highest final Mean Utility (0.809; 95\% task-bootstrap CI
$[0.698, 0.909]$) and Joint Pass rate (0.617). Its Mean Utility increases by
$0.156$ from Round 1 to Round 10, exceeding the gains of Full History
($+0.067$) and Random Replacement ($+0.086$), while Append-only decreases by
$0.025$. Second, TIDE obtains the best utility with only 2,121 active-memory
tokens, reducing memory usage by 20.1\% relative to Full History (2,653
tokens) and by 20.4\% relative to Append-only (2,666 tokens). Its performance
gain therefore does not depend on unbounded accumulation of historical
feedback.

Finally, lifecycle telemetry shows that 9 of TIDE's 10 \textsc{Evict}
operations target injected noisy memories (90.0\%), substantially exceeding
the 29 of 120 evictions for Random Replacement (24.2\%). The difference is
significant under a one-sided Fisher exact test
($p=5.81\times10^{-5}$). This result indicates that TIDE not only maintains a
bounded memory but also uses feedback credit to preferentially identify and
remove low-quality memories instead of relying on infrequent random deletion.
Overall, TIDE achieves higher final utility under a limited memory budget and
exhibits selective memory governance against injected noise.

\FloatBarrier
\section{Credit and Memory-Operator Ablations}
\suppressfloats[t]
\label{app:e2_credit_operator_ablation}
\paragraph{Experimental setup.}
E2 identifies whether TIDE's gains arise from temporal, semantic, and
responsibility credit or from the \textsc{Revise}, \textsc{Merge}, and
\textsc{Evict} operators, thereby opening the system beyond a black-box
comparison. We compare Full TIDE with three credit ablations that remove
temporal, semantic, or responsibility credit and three operator ablations that
remove \textsc{Revise}, \textsc{Merge}, or \textsc{Evict}. In our
implementation, \textsc{Revise} corresponds to feedback-directed mutation of
an individual memory, \textsc{Merge} implements the crossover or consolidation
of complementary memories, and \textsc{Evict} removes low-fitness memories
during capacity-constrained selection. For the w/o \textsc{Evict} variant,
once the memory reaches its capacity, newly generated candidates are not
admitted and no incumbent memory is removed. This preserves the common
capacity constraint without introducing an alternative eviction policy. All
seven conditions use the same initial memory, feedback order, memory capacity,
12 tasks, and seed 13. Each condition processes the same 12 candidate sets,
yielding 84 batches in total. We also keep the generation backbone, decoding
parameters, 2,000-token limit, and frozen judge fixed, while implementing an
explicitly distinct credit or editing policy for each ablation. The primary
metric is the task-level change in Mean Utility relative to Full TIDE. We
additionally report Mean Utility, Compliance, Format, Efficiency, batch-level
Diversity, and Joint Pass. Confidence intervals are computed using a paired
task bootstrap over the 12 tasks; a component contribution is treated as
detectable when the confidence interval of its difference excludes zero.

\paragraph{Results.}
\begin{figure}[htbp]
  \centering
  \includegraphics[width=\columnwidth]{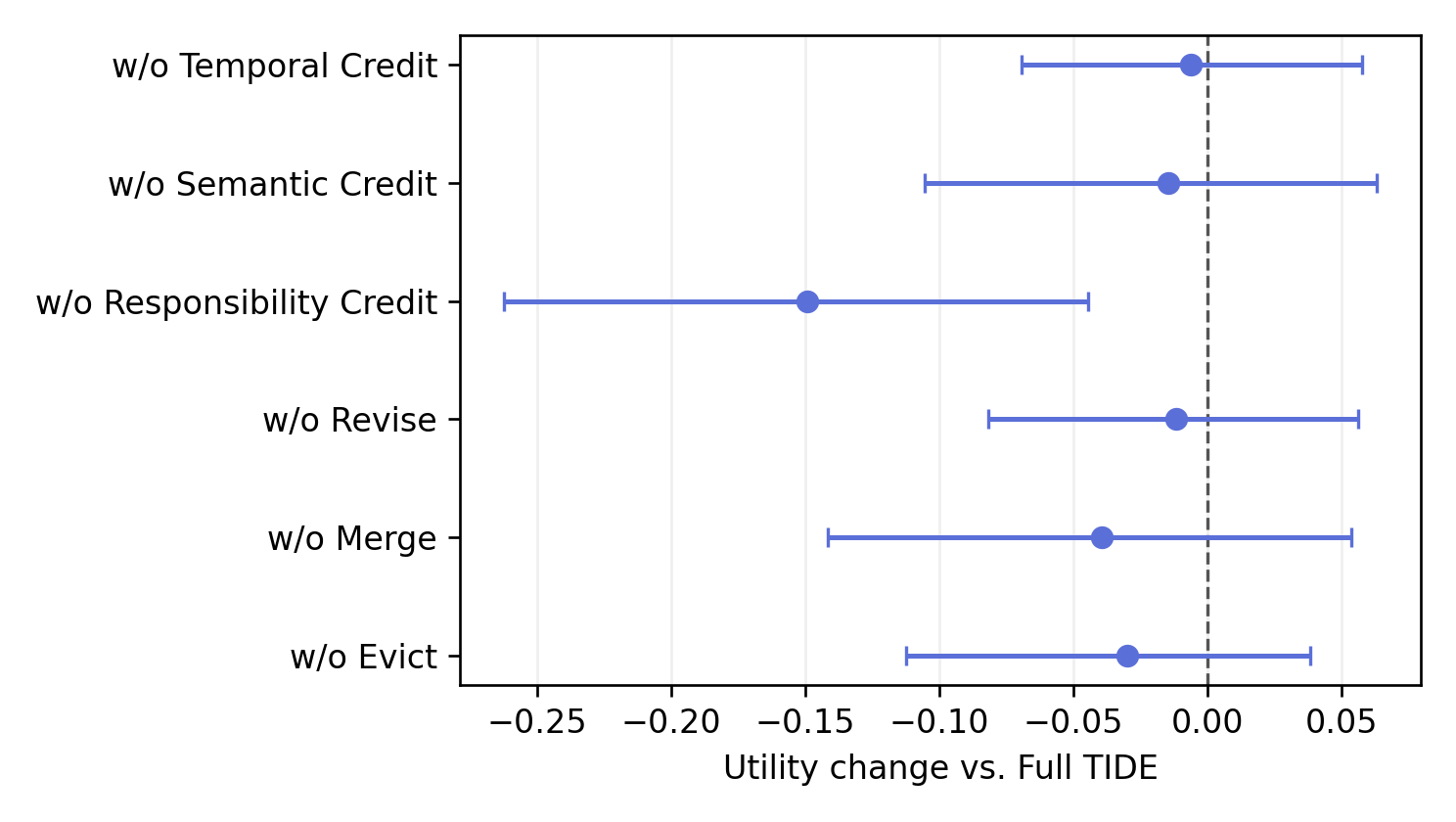}
  \caption{Credit and memory-operator ablations on 12 fixed tasks with seed
  13. Points show the paired task-level change in normalized Mean Utility
  relative to Full TIDE; horizontal bars denote task-bootstrap 95\% confidence
  intervals. Negative values indicate degradation. Responsibility credit is
  the only ablation whose interval excludes zero.}
  \label{fig:e2_credit_operator_ablation}
\end{figure}

\begin{table*}[htbp]
  \centering
  \caption{Credit and memory-operator ablations. Differences and confidence
  intervals are computed relative to Full TIDE using a paired bootstrap over
  the same 12 tasks.}
  \label{tab:e2_credit_operator_ablation}
  \small
  \setlength{\tabcolsep}{6pt}
  \renewcommand{\arraystretch}{1.08}
  \begin{tabular}{lccccc}
    \toprule
    Variant &
    \shortstack{Mean\\Utility $\uparrow$} &
    $\Delta$ vs. Full $\uparrow$ &
    95\% CI of $\Delta$ &
    \shortstack{Joint\\Pass $\uparrow$} &
    Diversity $\uparrow$ \\
    \midrule
    \textbf{Full TIDE}               & \textbf{0.852} & 0.000 & [0.000, 0.000] & 0.658 & 0.562 \\
    w/o Temporal Credit              & 0.846 & -0.006 & [-0.069, 0.058] & \textbf{0.725} & \textbf{0.708} \\
    w/o Semantic Credit              & 0.838 & -0.015 & [-0.106, 0.063] & 0.650 & 0.615 \\
    w/o Responsibility Credit        & 0.703 & \textbf{-0.149} & \textbf{[-0.262, -0.044]} & 0.292 & 0.469 \\
    w/o \textsc{Revise}              & 0.840 & -0.012 & [-0.082, 0.056] & 0.667 & 0.677 \\
    w/o \textsc{Merge}               & 0.813 & -0.040 & [-0.142, 0.053] & 0.617 & 0.667 \\
    w/o \textsc{Evict}               & 0.822 & -0.030 & [-0.112, 0.038] & 0.608 & 0.583 \\
    \bottomrule
  \end{tabular}
\end{table*}

The ablation results reveal the key mechanism behind TIDE's performance
advantage. Removing responsibility credit reduces Mean Utility from 0.852 to
0.703, corresponding to a paired difference of $-0.149$ (95\% CI
$[-0.262,-0.044]$), while Joint Pass drops sharply from 0.658 to 0.292. This
is the largest degradation among all ablations and the only one whose
confidence interval excludes zero, demonstrating that accurately attributing
feedback to the memories responsible for generation is central to effective
memory evolution in TIDE. Removing temporal credit, semantic credit,
\textsc{Revise}, \textsc{Merge},
or \textsc{Evict} also reduces Mean Utility by 0.006, 0.015, 0.012, 0.040,
and 0.030, respectively, indicating that the credit signals and editing
operators jointly support the complete system. Overall, Full TIDE achieves the
highest Mean Utility (0.852), validating the importance of combining
responsibility attribution with multiple memory-evolution operators to improve
aggregate generation quality.

\FloatBarrier
\section{Fitness--Diversity Dynamics and Population Health}
\suppressfloats[t]
\label{app:e3_population_health}

\paragraph{Experimental setup.}
E3 examines whether improvements in future utility come at the cost of memory
duplication or capacity expansion, thereby distinguishing high-quality
convergence from premature convergence. We reuse the four methods evaluated in
E1 and their 11 snapshots from Rounds 0--10, yielding 44 method--round
observations without additional material generation or judge calls. Semantic
distances are computed using embeddings from the frozen SFT Qwen3-4B model,
with the same model version, normalization procedure, and near-duplicate
threshold for all methods. We report Mean Utility, embedding-based Semantic
Diversity, Near-duplicate Rate, Effective Population Size, active population
size, Memory Churn, and Memory Tokens, and assess functional diversity jointly
with the frozen judge's batch-level Diversity score. Because short marketing
materials are tightly constrained by length, benefit terminology, channel
format, and target audience, embedding distance serves only as a conservative
diagnostic of surface-level semantic dispersion and should not be equated in
isolation with effective strategy diversity.

\paragraph{Results.}
\begin{figure*}[htbp]
  \centering
  \includegraphics[width=0.82\textwidth]{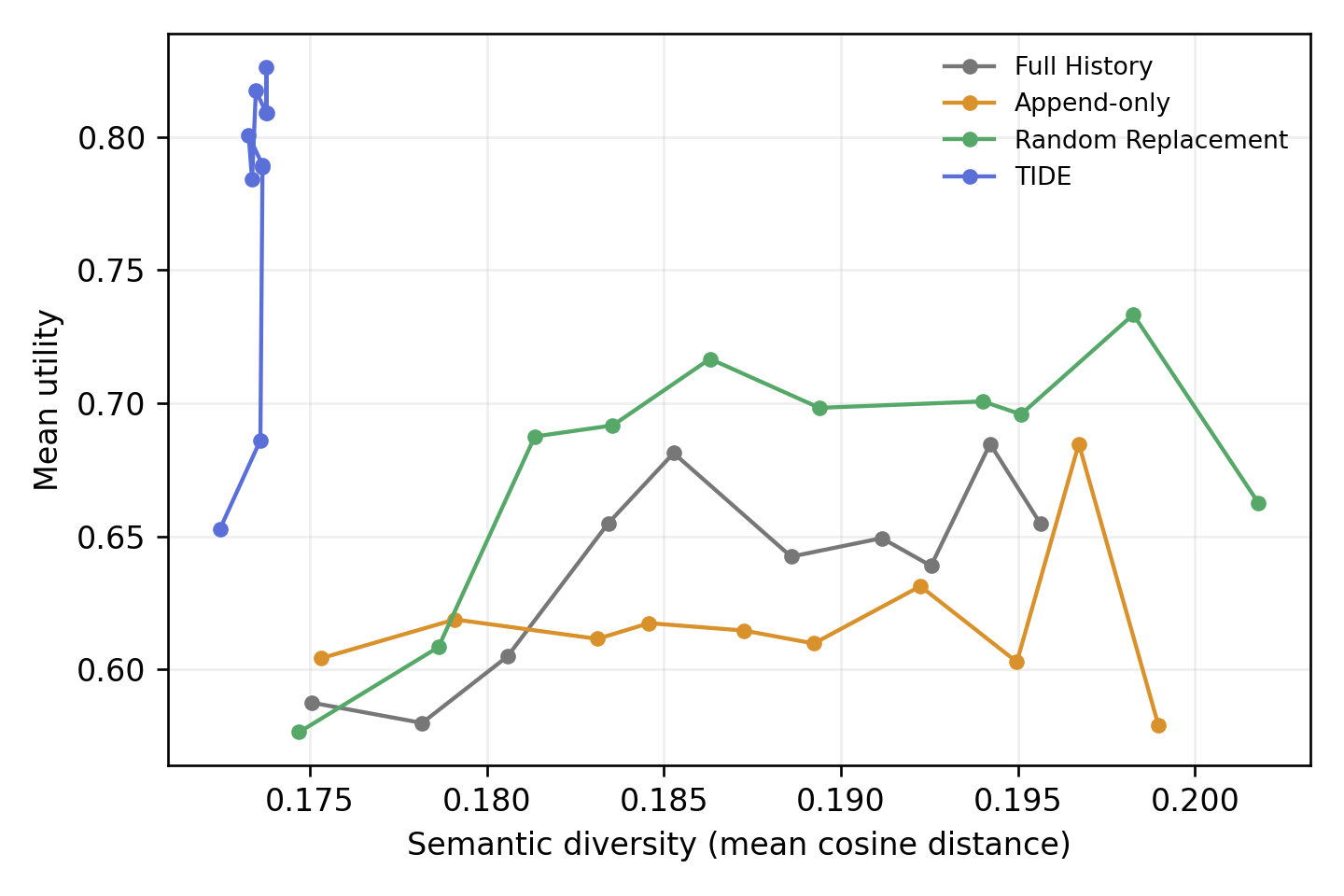}
  \caption{Quality--diversity trajectories over 11 memory snapshots on 12
  fixed tasks with seed 13. Color indicates the replay round, and marker size
  indicates active-memory tokens. TIDE attains the highest final utility with
  fewer memory tokens. Its lower embedding-space dispersion is interpreted
  jointly with Effective Population Size and batch-level Diversity because the
  feasible language space is tightly constrained.}
  \label{fig:e3_quality_diversity}
\end{figure*}

\begin{figure*}[htbp]
  \centering
  \includegraphics[width=0.82\textwidth]{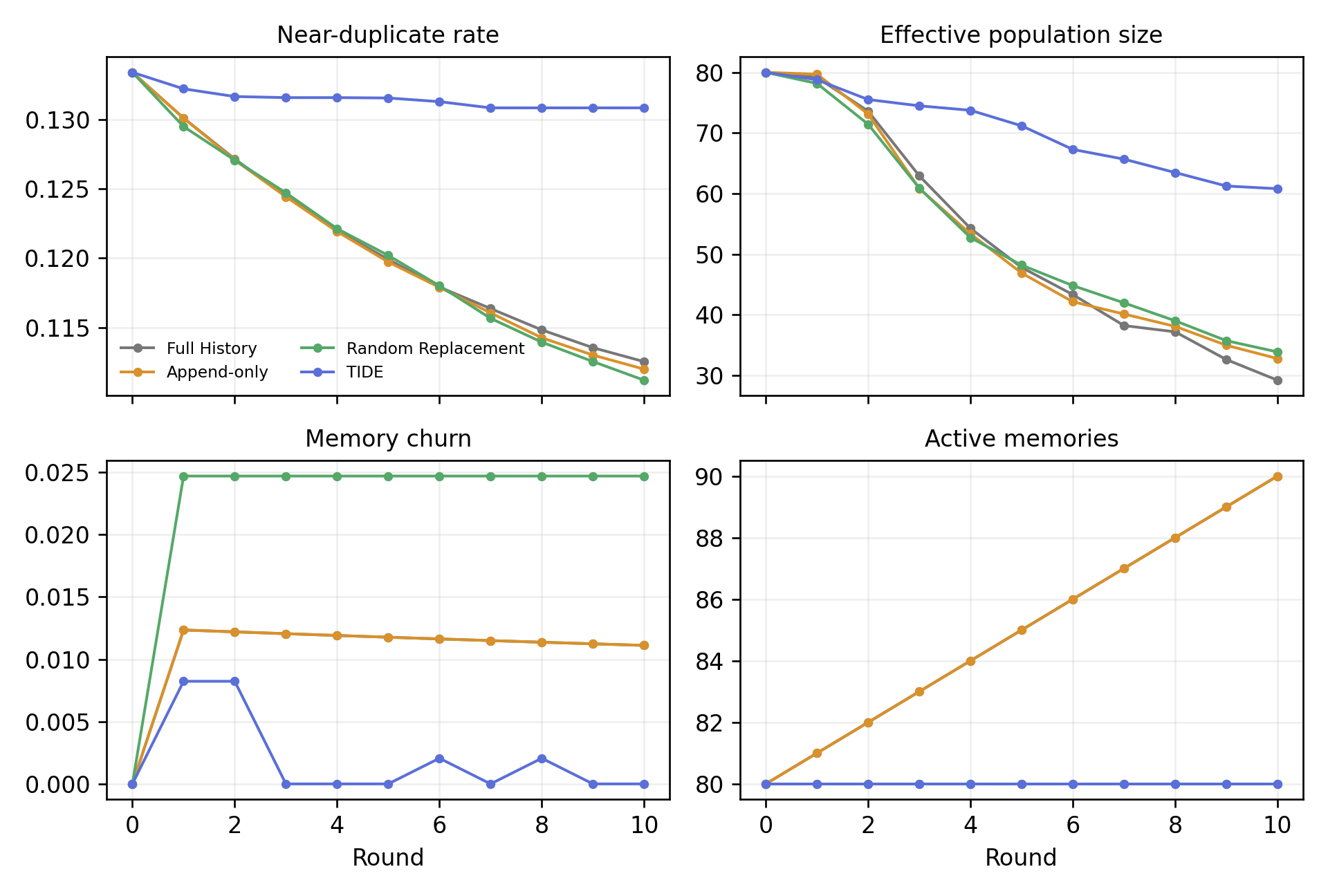}
  \caption{Population-health diagnostics across replay rounds. We report
  Near-duplicate Rate, Effective Population Size, active population size, and
  Memory Churn. TIDE keeps its active population bounded at 80 items and
  retains a substantially larger effective population than the baselines at
  Round 10 without unbounded token growth.}
  \label{fig:e3_population_health}
\end{figure*}

\begin{table*}[htbp]
  \centering
  \caption{Round-10 population quality and health. Utility and population
  statistics are averaged over the same 12 fixed tasks. Semantic Diversity
  and Near-duplicate Rate measure surface-level embedding dispersion within a
  tightly constrained material space; they are diagnostics rather than
  complete measures of functional diversity.}
  \label{tab:e3_population_health}
  \small
  \setlength{\tabcolsep}{5pt}
  \renewcommand{\arraystretch}{1.08}
  \begin{tabular}{lcccccc}
    \toprule
    Method &
    \shortstack{Mean\\Utility $\uparrow$} &
    \shortstack{Semantic\\Diversity $\uparrow$} &
    \shortstack{Near-duplicate\\Rate $\downarrow$} &
    \shortstack{Effective\\Population $\uparrow$} &
    \shortstack{Active\\Items} &
    \shortstack{Memory\\Tokens $\downarrow$} \\
    \midrule
    Full History       & 0.655 & 0.196 & 0.113 & 29.19 & 90 & 2,653 \\
    Append-only        & 0.579 & 0.199 & 0.112 & 32.77 & 90 & 2,666 \\
    Random Replacement & 0.663 & \textbf{0.202} & \textbf{0.111} & 33.86 & 80 & 2,423 \\
    \textbf{TIDE}      & \textbf{0.809} & 0.174 & 0.131 & \textbf{60.80} & 80 & \textbf{2,121} \\
    \bottomrule
  \end{tabular}
\end{table*}

At Round 10, TIDE simultaneously demonstrates higher generation quality, lower
memory cost, and a healthier effective population. Its Mean Utility reaches
\textbf{0.809}, clearly exceeding Full History (0.655), Append-only (0.579),
and Random Replacement (0.663); meanwhile, TIDE keeps its active memory bounded
at 80 items with an average cost of only \textbf{2,121 tokens}, showing that
its quality advantage does not depend on continually expanding memory
capacity. TIDE achieves an Effective Population Size of \textbf{60.80},
approximately 1.8--2.1 times those of the three baselines (29.19--33.86),
indicating that effective memory weight remains broadly distributed across
useful strategies rather than collapsing onto a few high-weight items.
Although its embedding-based Semantic Diversity is 0.174 and its
Near-duplicate Rate is 0.131, the frozen judge assigns a Round-10 batch-level
Diversity score of 0.500, comparable to Full History (0.510) and Append-only
(0.490) and higher than Random Replacement (0.438). Taken together, the larger
effective population and batch-level Diversity show that TIDE preserves
functional diversity comparable to the strongest baseline and higher than the
remaining bounded and append-only baselines, while allocating its limited
memory budget more effectively to efficient and compliant generation
strategies. The lower embedding dispersion primarily reflects surface-form
convergence under the length, benefit, channel, and audience constraints of
short marketing materials rather than a collapse of the effective strategy
population.

\FloatBarrier
\section{Adaptation and Recovery after a Distribution Shift}
\suppressfloats[t]
\label{app:e4_distribution_shift}

\paragraph{Experimental setup.}
E4 examines whether TIDE can adapt to a new distribution and recover future-task utility faster than Frozen Memory, Full History, and Append-only after the environment changes. Using 12 stratified tasks, seed 13, and a multidimensional scoring protocol,
we construct an event-aligned controlled shift: we construct an event-aligned controlled shift: materials are generated and updated under task distribution A before the change, after which the channel--benefit composition switches to task distribution B. Feedback continues to derive exclusively from the four evaluation dimensions and does not use real CTR/CVR or historical lifecycle labels. We compare Frozen Memory, Full History, Append-only, Random Replacement, and TIDE, and additionally evaluate TIDE with one- and two-round feedback delays. The main trajectories contain $5\times9\times12=540$ candidate sets, while the two delay conditions add 216 sets, yielding 756 sets in total. The experiment focuses on adaptation under a controlled distribution shift. We report Immediate Drop, Recovery Time, Post-shift Utility AUC, Post-shift MEG, Negative Transfer, Stale-memory Survival, and update and inference token consumption.

\paragraph{Results.}
\begin{figure*}[htbp]
  \centering
  \includegraphics[width=0.82\textwidth]{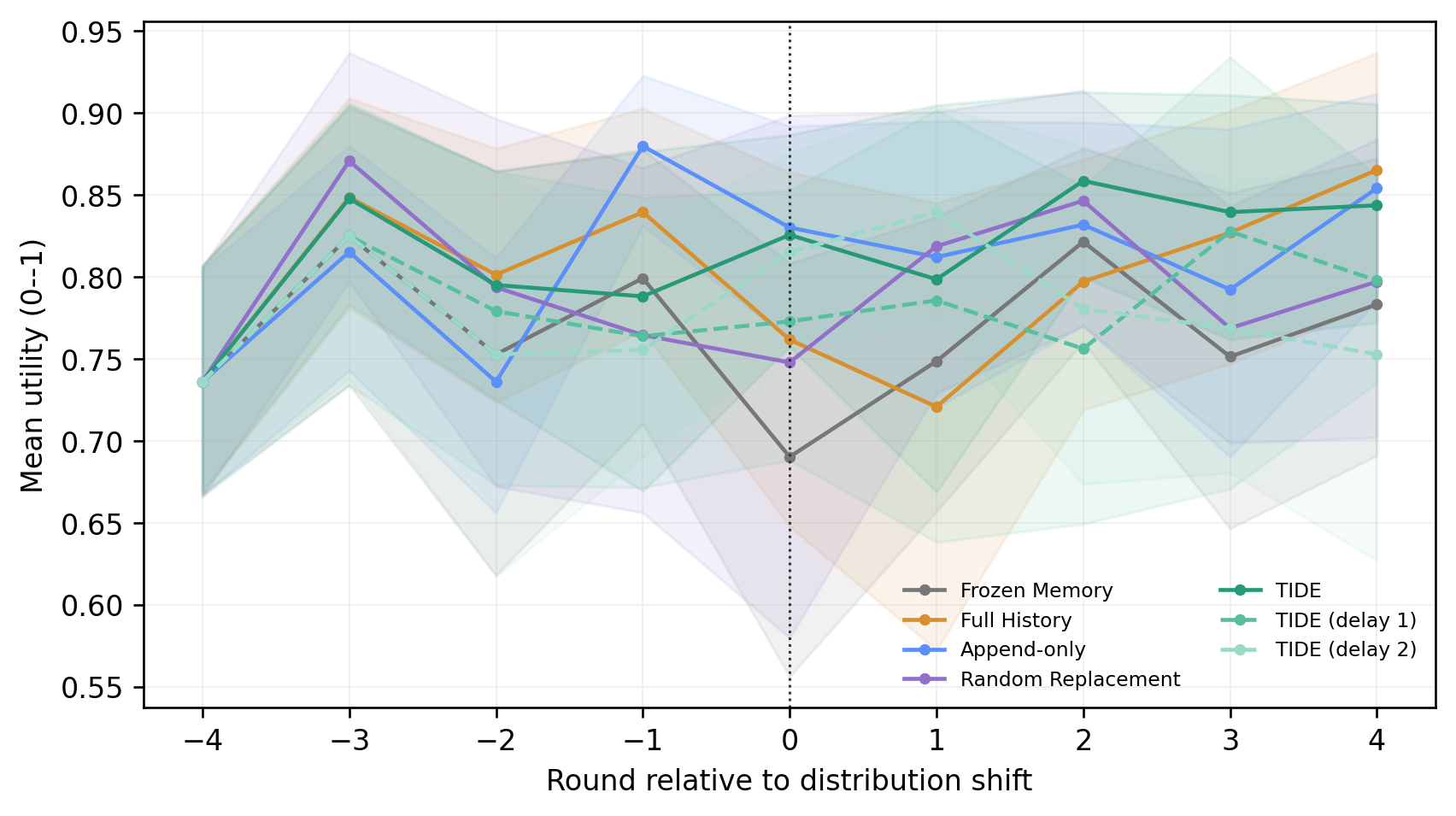}
  \caption{Recovery after a controlled distribution shift. Round 0 denotes the change point; curves show normalized future utility with task-level bootstrap 95\% confidence intervals. Results use 12 fixed tasks and seed 13.}
  \label{fig:e4_shift_recovery}
\end{figure*}

\begin{figure}[htbp]
  \centering
  \includegraphics[width=0.82\textwidth]{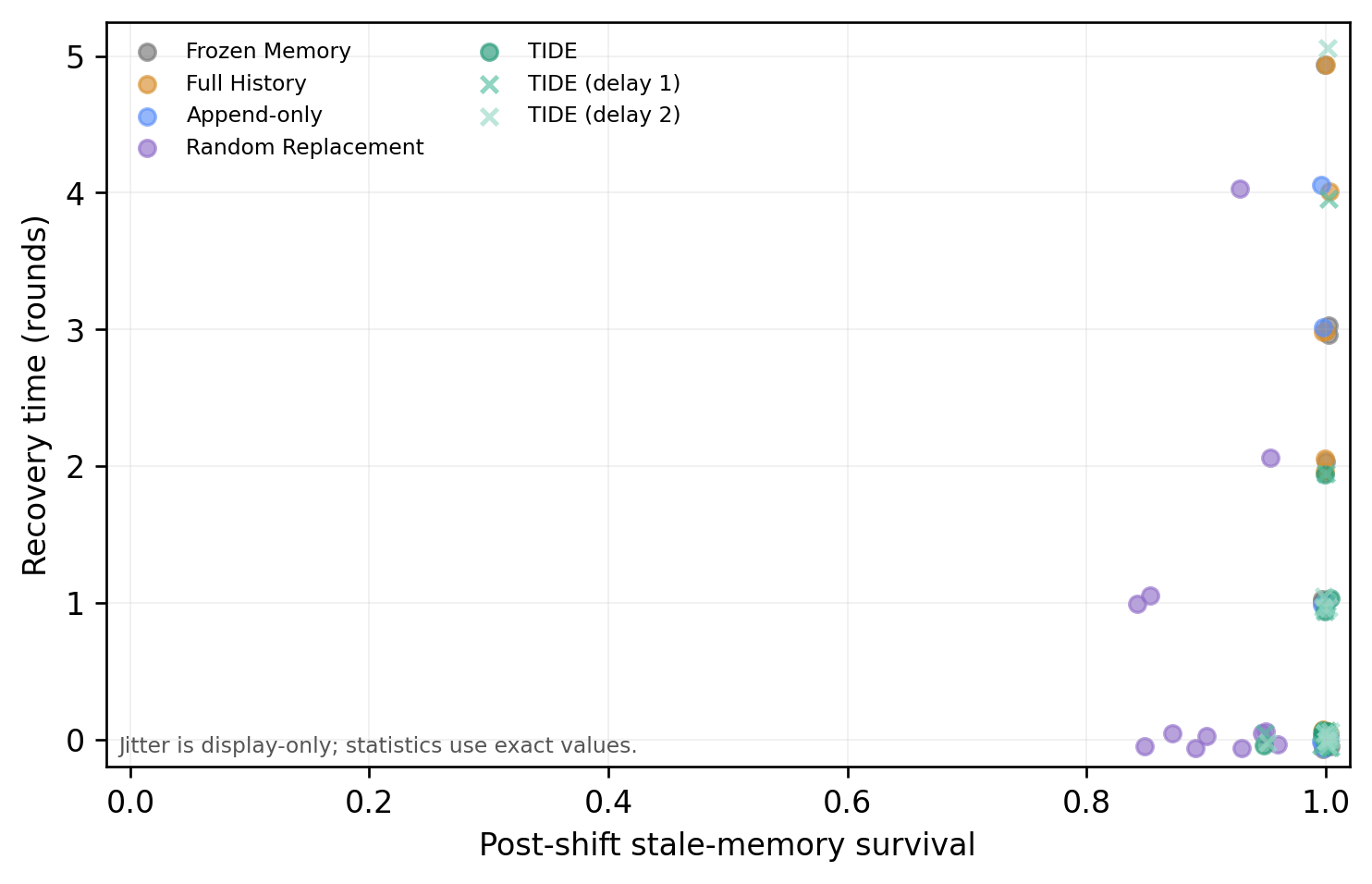}
  \caption{Task-level relationship between post-shift stale-memory survival and recovery time. Small deterministic jitter is used only to reveal overlapping observations; all reported statistics use the exact values. TIDE achieves rapid recovery despite high stale-memory survival, indicating that its adaptation advantage does not depend on aggressive memory deletion.}
  \label{fig:e4_stale_memory_recovery}
\end{figure}

\begin{table*}[htbp]
  \centering
  \caption{Adaptation and recovery after the controlled distribution shift. Recovery Time measures the number of rounds required to return to 95\% of pre-shift utility, and Post-shift MEG is measured relative to Frozen Memory.}
  \label{tab:e4_shift_recovery}
  \small
  \setlength{\tabcolsep}{6pt}
  \renewcommand{\arraystretch}{1.08}
  \begin{tabular}{lccccc}
    \toprule
    Method &
    \shortstack{Immediate\\Drop $\downarrow$} &
    \shortstack{Recovery\\Time $\downarrow$} &
    \shortstack{Post-shift\\AUC $\uparrow$} &
    \shortstack{Post-shift\\MEG $\uparrow$} &
    \shortstack{Stale-memory\\Survival $\downarrow$} \\
    \midrule
    Frozen Memory      & 0.088 & 1.25 & 0.759 & +0.000 & 1.000 \\
    Full History       & 0.045 & 1.58 & 0.794 & +0.035 & 1.000 \\
    Append-only        & 0.038 & 0.75 & 0.824 & +0.065 & 1.000 \\
    Random Replacement & 0.043 & 0.67 & 0.796 & +0.037 & 0.906 \\
    \textbf{TIDE}      & \textbf{0.034} & \textbf{0.33} & \textbf{0.833} & \textbf{+0.074} & 0.996 \\
    \bottomrule
  \end{tabular}
\end{table*}

After the distribution shift, TIDE demonstrates consistent advantages in adaptation quality, recovery speed, and cross-task stability. Its Post-shift Utility AUC reaches \textbf{0.833}, the highest among all methods, and its mean MEG relative to Frozen Memory is \textbf{$+0.074$} (task-level bootstrap 95\% CI $[0.024,0.121]$), showing a reliable post-shift utility gain. TIDE requires only \textbf{0.33 rounds} on average to recover to 95\% of its pre-shift utility, substantially faster than Random Replacement (0.67 rounds), Append-only (0.75 rounds), Frozen Memory (1.25 rounds), and Full History (1.58 rounds); its Immediate Drop of \textbf{$0.034$} is the smallest among all methods,
indicating that TIDE experiences the mildest average utility degradation at
the change point. Task-level paired results further show that TIDE achieves a higher post-shift AUC than Frozen Memory, Full History, Append-only, and Random Replacement on 9/12, 8/12, 7/12, and 8/12 tasks, respectively, demonstrating that its advantage spans most tasks rather than being driven by a small subset. Even with feedback delayed by one or two rounds, TIDE retains Post-shift AUC values of 0.788 and 0.791, both above the 0.759 achieved by Frozen Memory, indicating resilience to delayed feedback. TIDE also achieves the fastest recovery while retaining a stale-memory survival rate of 0.996, showing that its adaptation advantage does not require large-scale deletion of historical memory. Overall, TIDE rapidly incorporates new feedback, restores future-task utility within a limited number of update rounds, and maintains the strongest average performance under the controlled distribution shift.

\begin{figure*}[htbp]
  \centering
  \includegraphics[width=1.0\textwidth]{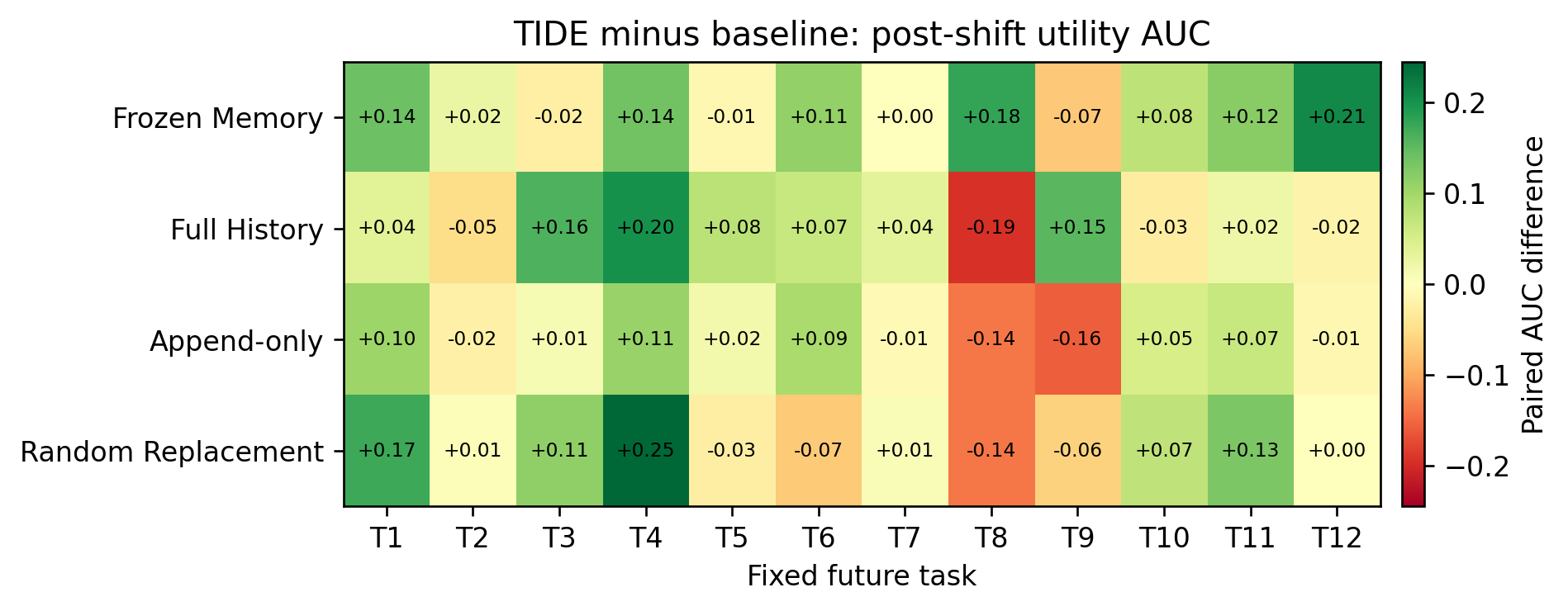}
  \caption{Task-level post-shift utility-AUC difference between TIDE and each baseline. Positive cells favor TIDE. Task aliases T1--T12 preserve the fixed paired evaluation units without exposing scene identifiers.}
  \label{fig:e4_taskwise_auc_advantage}
\end{figure*}

\begin{figure}[htbp]
  \centering
  \includegraphics[width=1.0\textwidth]{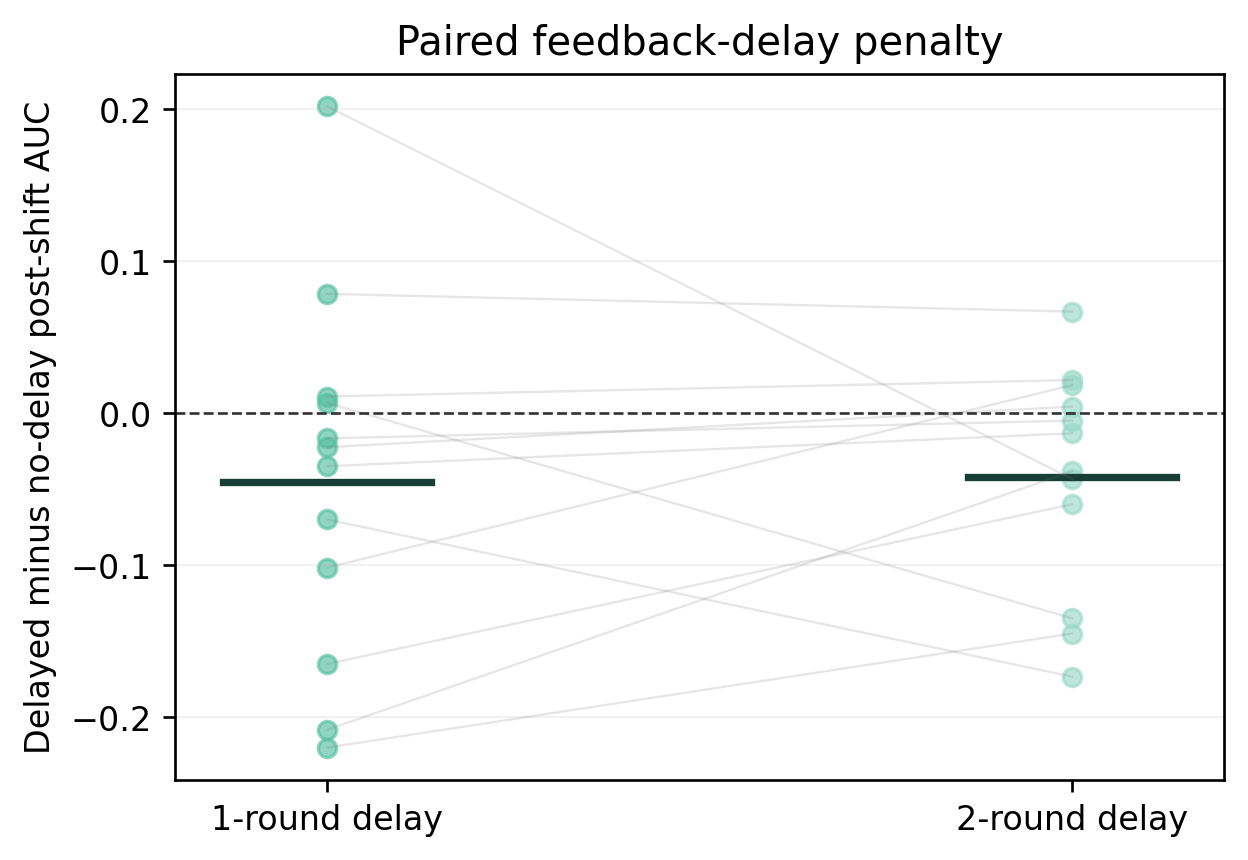}
  \caption{Paired effect of delaying TIDE feedback by one or two rounds. Each point represents a fixed future task; horizontal segments denote means across tasks. Values below zero indicate lower Post-shift AUC than under the no-delay TIDE condition.}
  \label{fig:e4_delay_sensitivity}
\end{figure}

\clearpage
\section{Memory-Operator Prompt Templates}
\suppressfloats[t]
\label{app:memory_operator_prompts}

We provide the prompt templates used by TIDE's four memory operators.
Bracketed fields denote runtime inputs. \textsc{Merge} implements crossover,
whereas \textsc{Revise} implements feedback-directed mutation. All generated
candidates retain their source-memory identifiers, evidence links, and version
information.

\begin{operatorbox}{Operator 1: Reinforce}
\textbf{Input:}
\texttt{\detokenize{[MEMORY]}},
\texttt{\detokenize{[CONTEXT]}},
\texttt{\detokenize{[ATTRIBUTED_EVIDENCE]}}, and
\texttt{\detokenize{[FITNESS_VECTOR]}}.

\textbf{Prompt:}
Determine whether the newly attributed evidence supports the existing memory
under the specified context. Preserve the original meaning and memory
identifier. Add only conclusions directly supported by the evidence, and
record the audience, channel, placement, campaign stage, and other conditions
under which the evidence is valid. Do not generalize beyond the observed
context. If the evidence contradicts the memory or remains insufficient,
return \texttt{not\_applicable} rather than reinforcing it.

\textbf{Output:}
decision, updated evidence, applicability conditions, confidence, and source
references.

\textbf{Safeguard:}
The reinforcement is committed only when the attributed evidence is valid,
contextually consistent, and traceable to its source records.
\end{operatorbox}

\begin{operatorbox}{Operator 2: Crossover (\textsc{Merge})}
\textbf{Input:}
\texttt{\detokenize{[PARENT_MEMORIES]}},
\texttt{\detokenize{[ATTRIBUTED_EVIDENCE]}}, and
\texttt{\detokenize{[TARGET_CONTEXT]}}.

\textbf{Prompt:}
Construct a candidate memory from the supplied parent memories. Identify
decisions that are compatible and complementary, while preserving their
context-specific conditions and exceptions. Resolve a conflict only when the
provided evidence supports a resolution. Do not merge memories that apply to
incompatible audiences, channels, placements, or campaign stages. Produce a
concise operational rule that provides additional reusable value rather than
simply concatenating the parent memories. Preserve the identifiers and
evidence links of all parents.

\textbf{Output:}
candidate content, applicability conditions, parent IDs, supporting sources,
resolved conflicts, unresolved conflicts, and confidence.

\textbf{Safeguard:}
The candidate is rejected when the parents are redundant, contextually
incompatible, unsupported by common evidence, or cannot be merged without
discarding an important exception.
\end{operatorbox}

\begin{operatorbox}{Operator 3: Mutation (\textsc{Revise})}
\textbf{Input:}
\texttt{\detokenize{[MEMORY]}},
\texttt{\detokenize{[CONTEXT]}},
\texttt{\detokenize{[ATTRIBUTED_EVIDENCE]}}, and
\texttt{\detokenize{[OBSERVED_FAILURE]}}.

\textbf{Prompt:}
Identify the smallest unsupported, incorrect, or overly general part of the
existing memory. Select one primary operation from retain, narrow, revise, or
split. Preserve all content that remains supported by the evidence, together
with its provenance. Introduce no conclusion that is not grounded in the
supplied feedback. When the evidence differs across contexts, prefer
narrowing or splitting the memory instead of globally overwriting it. Clearly
separate supported, unsupported, and unresolved applicability conditions.

\textbf{Output:}
operation, revised content, applicability conditions, preserved content,
removed or changed content, source-memory ID, supporting sources, and
confidence.

\textbf{Safeguard:}
A revision is committed only when it preserves the source-memory identifier
and complete evidence trace. Unsupported global rewrites are discarded.
\end{operatorbox}

\begin{operatorbox}{Operator 4: Evict}
\textbf{Input:}
\texttt{\detokenize{[MEMORY_RECORD]}},
\texttt{\detokenize{[FITNESS_EVIDENCE]}},
\texttt{\detokenize{[USAGE_HISTORY]}}, and
\texttt{\detokenize{[POPULATION_RELATIONS]}}.

\textbf{Prompt:}
Assess whether the candidate memory should remain in the
capacity-constrained population. Consider evidence reliability, contextual
usefulness, responsibility history, redundancy, conflict, and recency. Do not
remove a memory solely because it is old or infrequently retrieved. Prefer
eviction when the memory is repeatedly contradicted, superseded, redundant,
valid only in an obsolete context, or assigned persistently low fitness.
Protect a memory that covers a unique task or context unless strong
counterevidence exists. Use only feedback available before the current update
time.

\textbf{Output:}
decision, eviction score, reasons, supporting sources, protected unique
contexts, and confidence.

\textbf{Safeguard:}
The prompt produces an eviction assessment rather than directly deleting the
memory. Final removal is performed by
$\operatorname{GovernAndSelect}_B$ under the fixed capacity constraint, with
provenance and version history retained for audit.
\end{operatorbox}

\end{document}